\documentclass[journal]{IEEEtran}  

\IEEEoverridecommandlockouts                             
                                                         
\usepackage{graphicx} 
\usepackage{amsmath} 
\usepackage{amssymb}  
\usepackage{soul} 
\usepackage[normalem]{ulem} 
\usepackage{upgreek}

\usepackage{resizegather}

\usepackage{color}

\newtheorem{definition}{Definition}
\newtheorem{proposition}{Proposition}
\newtheorem{remark}{Remark}

\usepackage{hyperref} 
\hypersetup{ pdfborder = {0 0 .01} } 

\usepackage{bm}
\usepackage{comment}

\usepackage{cancel}

\usepackage{array}
\usepackage[]{algorithm2e}

\newcommand\numberthis{\addtocounter{equation}{1}\tag{\theequation}}

\newcommand{\Ro}{\mathbf{R}_o}
\newcommand{\To}{\mathbf{T}_o}
\newcommand{\Rh}{\mathbf{R}_h}
\newcommand{\Rf}{\mathbf{R}_{fi}}
\newcommand{\K}{\mathbf{K}_i}
\newcommand{\KOne}{\mathbf{K}_1}
\newcommand{\KTwo}{\mathbf{K}_2}
\newcommand{\KH}{\mathbf{K}^\mathcal{H}_i}
\newcommand{\KHone}{\mathbf{K}^\mathcal{H}_1}
\newcommand{\KHtwo}{\mathbf{K}^\mathcal{H}_2}
\newcommand{\DK}{\dot{\mathbf{K}}_i}
\newcommand{\fBt}{\mathbf{f}^\mathcal{B}_{ti}}

\newcommand{\ft}{\mathbf{f}_{ti}}
\newcommand{\Dft}{\dot{\mathbf{f}}_{ti}}
\newcommand{\fthat}{\hat{\mathbf{f}}_{ti}}
\newcommand{\fc}{\mathbf{f}_{ci}}
\newcommand{\fcAll}{\mathbf{f}_c}
\newcommand{\Dfc}{\dot{\mathbf{f}}_{ci}}
\newcommand{\fN}{\mathbf{f}_{Ni}}
\newcommand{\DfN}{\dot{\mathbf{f}}_{Ni}}

\newcommand{\di}{\mathbf{d}_i}
\newcommand{\pf}{\mathbf{p}_{fi}}
\newcommand{\pfOne}{\mathbf{p}_{f1}}
\newcommand{\pfTwo}{\mathbf{p}_{f2}}
\newcommand{\pfAll}{\mathbf{p}_{f}}
\newcommand{\Dpf}{\dot{\mathbf{p}}_{fi}}
\newcommand{\po}{\mathbf{p}_o}
\newcommand{\Dpo}{\dot{\mathbf{p}}_o}
\newcommand{\pa}{\mathbf{p}_{ai}}

\newcommand{\pHa}{\mathbf{p}^\mathcal{H}_{ai}}
\newcommand{\paAll}{\mathbf{p}_{a}}
\newcommand{\Dpa}{\dot{\mathbf{p}}_{ai}}

\newcommand{\pBf}{\mathbf{p}^\mathcal{B}_{fi}}

\newcommand{\pHf}{\mathbf{p}^\mathcal{H}_{fi}}
\newcommand{\pBfAll}{\mathbf{p}^\mathcal{B}_{f}}
\newcommand{\DpBf}{\dot{\mathbf{p}}^\mathcal{B}_{fi}}
\newcommand{\nhat}{\hat{\mathbf{n}}_i}
\newcommand{\Dnhat}{\dot{\hat{\mathbf{n}}}_i}

\newcommand{\pHzero}{\mathbf{p}^\mathcal{H}_{ai}}

\newcommand{\dzero}{\mathbf{d}_{0i}}

\newcommand{\gn}{\mathbf{g}_{ni}}
\newcommand{\gc}{\mathbf{g}_{ci}}
\newcommand{\gN}{\mathbf{g}_{Ni}}
\newcommand{\gt}{\mathbf{g}_{ti}}
\newcommand{\ct}{\mathbf{c}_{ti}}
\newcommand{\cN}{\mathbf{c}_{Ni}}
\newcommand{\cf}{\mathbf{c}_{fi}}
\newcommand{\cc}{\mathbf{c}_{ci}}

\newcommand{\lden}{\lambda_{\text{den},i}}

\newcommand{\hi}{\mathbf{h}_i}
\newcommand{\ai}{\mathbf{a}_i}

\newcommand{\myT}{T}

\newcommand{\glan}{\mathbf{g}_{\lambda_i}}
\newcommand{\glanT}{\mathbf{g}^{\myT}_{\lambda_i}}
\newcommand{\clan}{c_{\lambda_i}}
\newcommand{\glanInv}{\mathbf{g}_{\lambda_i}^\dagger}

\newcommand{\pe}{\mathbf{p}_{ej}}

\newcommand{\we}{\mathbf{w}_{ej}}
\newcommand{\weAll}{\mathbf{w}_e}

\newcommand{\wc}{\mathbf{w}_{ci}}
\newcommand{\wcAll}{\mathbf{w}_c}
\newcommand{\wkhat}{\mathbf{w}_{jk}}
\newcommand{\fkhat}{\hat{\mathbf{f}}_{jk}}
\newcommand{\What}{\mathbf{W}}
\newcommand{\WhatT}{\mathbf{W}^\myT}
\newcommand{\wg}{\mathbf{w}_g}
\newcommand{\dwc}{\delta\mathbf{w}_{c}}

\newcommand{\dwe}{\delta\mathbf{w}_{e}}
\newcommand{\wcbar}{\bar{\mathbf{w}}_c}

\newcommand{\dbeta}{\delta \bm \beta}

\newcommand{\ph}{\mathbf{p}_h}
\newcommand{\Dph}{\dot{\mathbf{p}}_h}

\newcommand{\BF}[1]{\bm{\mathsf{#1}}}

\newcommand{\B}{\mathcal{B}}
\newcommand{\W}{\mathcal{W}}
\newcommand{\HH}{\mathcal{H}}

\newcommand{\Wperp}{{}^{\perp}\mathbf{W}}

\newcommand{\fcperp}{{}^{\perp}\mathbf {\hat f}_{ci}}
\newcommand{\fcperpOne}{{}^{\perp}\mathbf {\hat f}_{c1}}
\newcommand{\fcperpTwo}{{}^{\perp}\mathbf {\hat f}_{c2}}

\newcommand{\WCe}{\mathcal{WC}_{e}}

\newcommand{\colwidth}{3.4in}
\newcommand{\real}{\mathbb{R}}

\title{In-hand Sliding Regrasp with Spring-Sliding Compliance}

\author{Jian Shi and Kevin M. Lynch 
  \thanks{Jian Shi is with Dorabot, Inc., Atlanta, GA USA (email: jian.shi@dorabot.com) }
  \thanks{Kevin M. Lynch is with the Center for Robotics and
    Biosystems and Mechanical Engineering Dept., Northwestern University, Evanston, IL 60208 USA. (email: kmlynch@northwestern.edu). He is also affiliated with the Northwestern Institute on Complex Systems (NICO).}
\thanks{This work was supported by NSF grant IIS\,-\,1527921.  We would like to thank Paul Umbanhowar, Zack Woodruff, and Nelson Rosa for their helpful suggestions and comments, and Huan Weng for his work on building the stiffness controller for the Allegro hand.}
}

\begin{document}

\maketitle
\begin{abstract}
We investigate in-hand regrasping by pushing an object against an external constraint and allowing sliding at the fingertips.  Each fingertip is modeled as attached to a multidimensional spring mounted to a position-controlled anchor.  Spring compliance maps contact forces to spring compressions, ensuring the fingers remain in contact, and sliding ``compliance'' governs the relationship between sliding motions and tangential contact forces.  A spring-sliding compliant regrasp is achieved by controlling the finger anchor motions.
  
We derive the fingertip sliding mechanics for multifingered sliding regrasps and analyze robust regrasping conditions in the presence of finger contact wrench uncertainties.  The results are verified in simulation and experiment with a two-fingered sliding regrasp designed to maximize robustness of the operation.  
\end{abstract}

\section{Introduction}
\label{sec:intro}

In-hand manipulation, and specifically regrasping an object within the hand, offers the promise of increased manipulator dexterity~\cite{Shi2017,chavan-dafle2014}.  Regrasp can be achieved purely by forces applied by the fingers themselves, or it can be achieved by taking advantage of external forces on the object.  As one example, in our previous work regrasp is achieved by accelerating the object such that the inertial load can no longer be resisted by friction with the fingers, causing sliding of the object~\cite{Shi2017}.  Short bursts of such motion can be used to achieve controllable dynamic in-hand sliding regrasps.

\begin{figure}
  \centering
  \includegraphics[width=\colwidth]{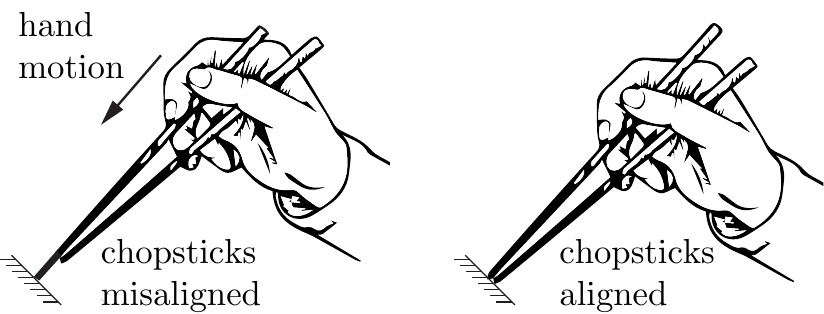}
  \caption{The extended (top) chopstick is pushed against a constraint, bringing it into alignment with the other chopstick by a sliding regrasp.}
  \label{fig:chopsticks}
\end{figure}

In this paper, we focus on quasistatic sliding regrasps taking advantage of contacts between the object and a rigid environment.  An example is shown in Figure~\ref{fig:chopsticks}.  After picking up a pair of chopsticks, often the ends of the chopsticks are misaligned, making the chopsticks difficult to use.  One strategy is to push the chopsticks against a constraint, bringing the ends into alignment.  During this operation, one (or both) of the chopsticks slides within the grasp.

We model each finger as a frictional point contact connected by a three-dimensional linear spring to an anchor point whose motion is controlled in three linear directions. Given the stiffness matrix governing the multidimensional spring, by position-controlling the anchor we can control the force applied to the object and initiate sliding when the contact force reaches the boundary of its friction cone. External contacts provide forces that maintain object force balance during the quasistatic sliding regrasp.

Similar to spring compliance that governs the relationship between contact forces and displacements, frictional sliding is a kind of nonlinear damping ``compliance'' that governs the relationship between tangential frictional forces and tangential sliding velocities.  Sliding compliance is a passive dissipative mechanical effect, requiring no active feedback control.
Spring-sliding compliance models are simple and compact (e.g., no finite-element elastic models) but can approximate many real-world contact interactions.  Features of this model of contact interaction include:

\begin{itemize}
	\item Spring compliance ensures that fingers remain in contact while sliding over general surfaces.  Spring compliance may be mechanically programmable and passive, ensuring stability~\cite{Hanafusa1977,Howard1996}.
	\item With spring compliance, contact forces are determined by finger compressions, so contact force control can be achieved by controlling finger anchor motions and sensing the compression.
	\item Sliding compliance bounds the possible tangential contact forces and allows sliding for in-hand regrasp.
\end{itemize}

\begin{figure}
	\centering
	\includegraphics[width = 3.4in]{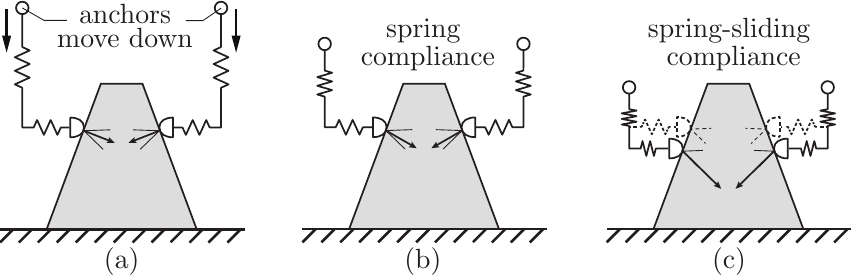}
	\vspace{-0.0in}
	\caption{Two springy fingers grasp a trapezoid with two point contacts. The spring anchors move down vertically and push the object against a fixed table. Lines at the contact points show friction cones and arrows show the contact forces applied to the object. When the anchors move from (a) to (b), the vertical springs are compressed and the contact forces increase in the vertical direction. The fingertips are still sticking since the contact forces are within the friction cones. As the anchors move to (c), the contact forces reach the edges of the friction cones and the fingertips start to slide on the contact surfaces. Contact forces from the table keep the object stationary and wrench-balanced. }
	\label{fig:CCexample}
\end{figure}

Figure~\ref{fig:CCexample} shows an example of an in-hand sliding regrasp of a trapezoid. When the anchors move down, at first the fingertips remain stationary, the finger springs compress, and the contact forces move toward the boundaries of the friction cone.  Once the contact forces reach the friction cone boundaries, the fingertips begin to slide and the springs continue to compress.  The grasped object is in quasistatic wrench balance if the sum of the gravitational wrench and contact wrench with the table balances the sum of the finger contact wrenches.

For a given object and rigid environment, we define a grasp configuration as the configuration of the object, the configuration of the fingertips relative to the object, and the configuration of the finger anchors.  The goal of this work is to design a quasistatically consistent (force-balanced at all times) set of finger anchor motions and the object motion relative to the rigid environment such that the fingertips achieve a desired new configuration relative to the object.  In the general formulation, the object could slide or roll at its contacts with the environment during the sliding regrasp, but in this paper we focus particularly on the case where the object remains stationary against the environment.  This allows us to design sliding regrasps that are robust to force disturbances, in a sense to be defined in Section~\ref{sec:robustness}.

After reviewing related work and the problem description, this paper has the following structure:
\begin{itemize}

\item \emph{Finger spring compliance model} (Section~\ref{sec:finger_compliance_model}):  This section describes finger designs and controls that fit the spring-compliance model.
  
	\item \emph{Finger contact mechanics} (Section~\ref{sec:contact_mechanics}): In this section we derive the mechanics of spring-sliding contact.  In particular, given a grasp configuration and the object's motion, this section derives the relationship between anchor velocities and fingertip velocities.  
	
	\item \emph{Object mechanics} (Section~\ref{sec:obj_mechanics}): This section describes the quasistatic wrench-balance conditions considering the object's motion
          and wrenches due to the external contacts, fingertips, and gravity.
	
	\item \emph{Robustness analysis} (Section~\ref{sec:robustness}): A planned sliding regrasp is robust to finger contact wrench uncertainty if the planned regrasp succeeds in the face of this uncertainty.

	\item \emph{Sliding regrasp planning} (Section~\ref{sec:motion_planning}): This section describes a general approach to finding feasible and robust object and finger anchor trajectories that realize a desired regrasp.

	\item \emph{Implementation} (Section~\ref{sec:implementation}): We describe a particular spring-sliding regrasp motion planner for the case of a two-fingered regrasp, where the objective is to maximize robustness of the spring-sliding regrasp.  Simulation and experimental results validating the approach are given.
\end{itemize}

Section~\ref{sec:conclusion} concludes with directions for future research.

%
\section{Related Work}
\label{sec:related-work}

\newcommand{\etal}{et al.}

\subsection{In-hand Manipulation}
As described in pioneering early work, in-hand manipulation involves adjusting finger contacts relative to an object using rolling~\cite{fearing1986,Cole1989,Cherif1998}, gaiting~\cite{Rus1999}, or sliding~\cite{cole1992}. Li et al.~\cite{li1989} and Yoshikawa and Nagai \cite{yoshikawa1991} used rigid, rolling finger contacts to calculate grasp stability, manipulability, and to develop controllers for tracking a position trajectory while maintaining a desired grasp force. Trinkle and Hunter extended the dexterous manipulation planning problem to consider rolling and slipping contact modes~\cite{trinkle1991}. The hybrid planning problem was further developed by Yashima et al.~\cite{yashima2003}. 
Brock addressed the problem of controlled in-hand sliding by first generating a constraint state map which outlines constraints on a grasped object due to the contact types and forces~\cite{brock1988}. By varying contact forces, controlled sliding was achieved in desired directions for a grasped cylinder. Sundaralingam and Hermans demonstrated in-hand rolling manipulation using only kinematic models~\cite{Sundaralingam2018}. To address inevitable errors or uncertainties in purely model-based approaches, iterative learning control~\cite{Yashima2018} and model-based reinforcement learning~\cite{Kumar2016} have been applied to learn a specific in-hand manipulation task over a series of trials.

Expanding in-hand manipulation to include dynamics, Furukawa et al.~demonstrated regrasping by tossing a foam cylinder and catching it~\cite{furukawa2006}. 
Chavan-Dafle et al.~tested hand-coded regrasps that take advantage of
external forces such as gravity, dynamic forces, and contact with the
environment to regrasp objects using a simple manipulator
\cite{chavan-dafle2014}.  Hou et al.\ studied dynamic planar pivoting
of a pinched object driven by hand swing motion and contact normal
force control~\cite{yifan2016}.  Vi\~{n}a et~al.\ showed that by using
adaptive control with vision and tactile feedback, monodirectional
pivoting of an object pinched by a pair of fingers can be achieved by
changing the gripping forces~\cite{vina2016}. 
Cruciani et al.\ derived a Dexterous Manipulation Graph to plan paths for a parallel-jaw gripper to slide along parallel surfaces of an object from one stable grasp to another~\cite{Cruciani2018}.
Sintov and Shapiro developed an algorithm to swing up a rod by generating gripper motions, where the contact point was modeled as a pivot joint that can apply frictional torques~\cite{Sintov2016}.  In our prior work, we used inertial loads to achieve in-hand sliding regrasps~\cite{Shi2017}.

Chavan-Dafle et al. explored in-hand manipulation of an object by external contacts with environmental constraints, as in this paper~\cite{chavan-dafle2015,chavan-dafle2018,Chavan-Dafle2018b}.  A laminar object is squeezed between two fingers and pushed against a constraint to cause sliding at the fingers.  They showed that such actions are similar to pushing an object sliding on a planar surface~\cite{Lynch96c}, and that sequences of pushes can be planned to achieve an in-hand regrasp.  In this paper, we explicitly model spring compliance so that in-hand sliding regrasp is possible with more complex grasp configurations, where the object is not laminar and any number of fingers can be in contact.

In recent work, Dollar et al.\ demonstrated a simple and robust type of in-hand manipulation based on the clever use of fingers that can switch between two different friction coefficients: high, for rolling or sticking contact, and low, for sliding manipulation~\cite{Dollar2019}. A laminar object, such as a square, is supported by a table and manipulated in the plane by two flat one-joint fingers.  Depending on the friction coefficient employed at each finger, the object can be made to slide or roll in the two-finger hand, achieving in-hand manipulation.

In this paper, the mechanics of spring-sliding compliance for in-hand regrasp are derived for generic 3D object geometries with no restriction on the number of fingers in contact.

\subsection{Compliant Grasps}

Spring-compliant grasps are a subset of spring-sliding-compliant grasps, as studied in this paper.
Hanafusa and Asada modeled the spring compliance of frictionless elastic fingers and formulated a notion of grasp stability~\cite{Hanafusa1977}. In their definition, a stable grasp means that the grasp restores the object to its initial configuration after a small configuration disturbance. Grasp stability is determined by finger stiffness and local contact geometry.
Baker \etal{}\ further developed the stability conditions under the same assumptions~\cite{Baker1985}. More generally, 
Howard and Kumar classified categories of equilibrium grasps and derived conditions for stability~\cite{Howard1996}.
Odhner and Dollar demonstrated in-hand rolling with an underactuated compliant hand~\cite{Odhner2015}.  
Cutkosky and Kao achieved a desired grasp stiffness by controlling finger joint stiffness~\cite{Cutkosky1989}.
Cutkosky and Kao also modeled sliding manipulation with spring compliance and limit surface frictional contacts~\cite{Kao1992}. The motions of the contact points were solved by assuming infinitesimal motions while the magnitude of the sliding velocity is fixed.
In this paper we allow finite sliding velocities and solve for the sliding velocity using the constraint that sliding contact forces are on the boundary of the friction cone. 

Spring-compliant grasps have applications in assembly. The remote center of compliance (RCC) device is a mechanical solution to reduce mating forces and the chance of jamming in certain assembly operations \cite{whitney1982}. Goswami and Peshkin generalized the idea by outlining a design strategy for passive devices to implement desired spring characteristics~\cite{Goswami1993}. Schimmels and Peshkin derived conditions for accommodation control to yield error-corrective assembly with frictional contacts~\cite{Schimmels1992, Schimmels1994}.
Ji and Xiao explored methods to plan compliant assembly based on a contact state graph~\cite{Ji2001}. Meeussen \etal\ developed an approach to convert a contact path into a force-based task specification for executing the compliant path via hybrid position and force control~\cite{Meeussen2005}.
Park \etal\ developed a procedure and a controller that yield compliant behavior using neither force feedback nor passive compliance mechanisms to solve the peg-in-hole assembly problem~\cite{Park2017}. 

\section{Problem Description}
\label{sec:definition}%
An $n$-fingered hand grasps an object with $n$ point contacts.
Each finger consists of an individually motion-controlled anchor point that is connected by a three-dimensional linear spring to a point fingertip.  
The object contacts a rigid stationary environment with a total of $m$ frictional point contacts.\footnote{A line contact is modeled by two point contacts and a face contact is modeled by three or more points.} 
A grasp configuration is defined by the positions of the finger anchors, the finger contact points, and the object's configuration. 
The problem can be described as: given (1) an initial grasp configuration where the object is in force balance and (2) a desired new grasp configuration, find quasistatically-consistent anchor and object motions that realize the regrasp.

\subsection{Assumptions}
\label{subsec:assumptions}

\begin{enumerate}
	\item Gravity and contact wrenches are always balanced (quasistatic assumption).
	\label{enum:quasistatic_assumption}
	\item Fingers contact the object at point fingertips.
	\label{enum:finger_assumption}  
      \item Each finger is linearly springy and the stiffness is known. Each $3\times 3$ stiffness matrix is symmetric and positive definite.
	\label{enum:stiff_assumption}
	\item Each finger maintains a positive contact normal force.
	\label{enum:fc_assumption}
	\item The object is rigid, smooth, and of known geometry.
	\item Dry Coulomb friction applies at each point contact.  During sliding contact, the tangential friction force $\mathbf{f}_t$ is aligned with the tangential sliding direction and has a magnitude $\mu f_N$, where $\mu \geq 0$ is the friction coefficient and $f_N > 0$ is the magnitude of the normal force; and during sticking contact, the total contact force is confined to a friction cone satisfying $\|\mathbf{f}_t\| \leq \mu f_N$.     The friction coefficients at all contacts are known, though this assumption is relaxed in our robustness analysis. For convenience, we assume that finger contacts with the object have a friction coefficient $\mu$ and environment contacts with the object have a friction coefficient $\mu_e$.
	\item The $m$ external contact points are known and the environment is assumed rigid and stationary.
	\label{enum:ex_contact_assumption}
\end{enumerate}

\begin{figure}
	\centering
	\includegraphics[width=3.4in]{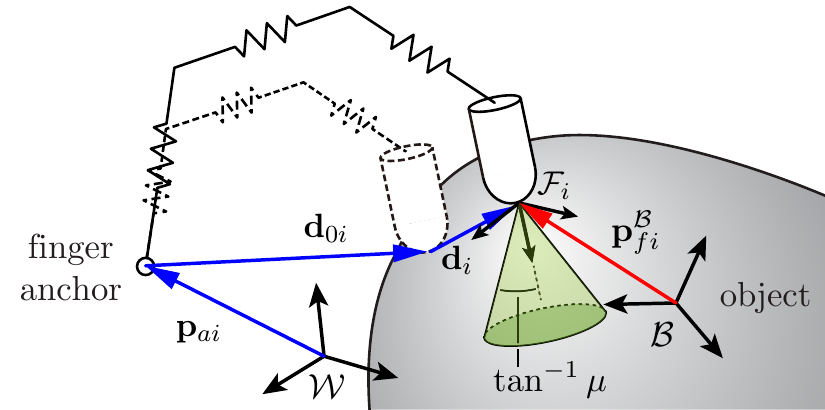}
	\vspace{-0.in}
	\caption{Finger notation.  The contact friction cone is indicated in green.}
	\label{fig:config}
\end{figure}

\subsection{Notation}
Vectors are written in bold lowercase letters, matrices are in bold capital letters, scalars are italicized, and coordinate frames are denoted with calligraphic letters. All variables are expressed in a world frame $\mathcal{W}$ unless noted otherwise in the superscripts. 
For example, $\pf$ is the fingertip position of the $i$th finger in the world frame $\W$ and $\pBf$ is the fingertip position in the object frame $\B$.  Frames of reference are typically chosen to simplify the mathematical expressions; standard transformations are used to move between frames.  Figure~\ref{fig:config} illustrates some of the quantities for a single finger.

\newcommand{\tw}{3in}
\newcounter{mycounter}  
\newenvironment{noindlist}
 {\begin{list}{\emph{\arabic{mycounter})}~~}{\usecounter{mycounter} \labelsep=0em \labelwidth=0em \leftmargin=0em \itemindent=0em}}
 {\end{list}}

 \vspace*{0.1in}
\subsubsection{Object Notation}~

\begin{tabular}{@{}p{0.375in}@{}p{3in}}
$\mathcal{B}$ & Frame attached to the object. \\
\end{tabular}

\begin{tabular}{@{}p{0.375in}@{}p{3in}}
$\mathbf{p}_o$ & The position of the origin of $\mathcal{B}$, $\po= [x_o, y_o, z_o]^\myT$.\\
\end{tabular}

\begin{tabular}{@{}p{0.375in}@{}p{3in}}
$\Ro$ & Rotation matrix representing the orientation of the object, $\Ro \in SO(3)$.\\
\end{tabular}

\begin{tabular}{@{}p{0.375in}@{}p{3in}}
$\To$ & Object configuration constructed of $\po$ and $\Ro$, $\To \in SE(3)$.\\
\end{tabular}

\begin{tabular}{@{}p{0.375in}@{}p{3in}}
$\bm \omega_o$ & Object angular velocity, $\bm \omega_o \in \real^3$.\\
\end{tabular}

\vspace*{0.1in}
\subsubsection{Finger Notation}~

\begin{tabular}{@{}p{0.375in}@{}p{3in}}
$\mathcal{F}_i$ & Finger frame attached to the $i$th ($i=1,...,n$) fingertip. The $z$-axis of $\mathcal{F}_i$ is aligned with the contact normal pointing into the object.\\
\end{tabular}

\begin{tabular}{@{}p{0.375in}@{}p{\tw}}
$\mathbf{p}_{fi}$ &  The $i$th fingertip position, $\pf = [x_{fi}, y_{fi}, z_{fi}]^\myT$.\\
\end{tabular}

\begin{tabular}{@{}p{0.375in}@{}p{\tw}}
$\Rf$ & Rotation matrix representing the orientation of $\mathcal{F}_i$. \\
\end{tabular}

\begin{tabular}{@{}p{0.375in}@{}p{\tw}}
$\mathbf{p}_{ai}$ & The $i$th anchor position, $\pa= [x_{ai}, y_{ai}, z_{ai}]^\myT$.\\
\end{tabular}

\begin{tabular}{@{}p{0.375in}@{}p{\tw}}
$\dzero$ & The equilibrium position of the $i$th fingertip.
\end{tabular}

\begin{tabular}{@{}p{0.375in}@{}p{\tw}}
$\mathbf{d}_i$ & Compression of the $i$th finger, $ \mathbf{d}_i= \pf - \dzero -\pa$.\\
\end{tabular}

\begin{tabular}{@{}p{0.375in}@{}p{\tw}}
  $\mathbf{K}_i$ & Stiffness matrix of the $i$th finger, $\mathbf{K}_i \in \mathbb{R}^{3 \times 3}$,
                   which may or may not depend on the finger joint configuration or other parameters.
\end{tabular}

\vspace*{0.1in}
\subsubsection{Contact Forces}

The contact force applied to the object by the $i$th finger is
\begin{equation}
	\mathbf{f}_{ci} = -\mathbf{K}_i \mathbf{d}_i = -\K(\pf - \pa - \dzero).
	\label{eq:fc}
\end{equation}
The contact normal into the object is a function of the finger contact position in $\B$,
\begin{equation}
	\hat{\mathbf{n}}_i(\pf^{\mathcal{B}}) = \mathbf{R}_{fi} [0, 0, 1]^\myT,
	\label{eq:n^hat}
\end{equation}
where the hat means the vector is a unit vector.
The contact normal force is the projection of $\fc$ to the normal direction,
\begin{equation}
	\mathbf{f}_{Ni} = (\mathbf{f}_{ci} \cdot \hat{\mathbf{n}}_i)\hat{\mathbf{n}}_i = \fc^\myT \nhat \nhat,
	\label{eq:fN}
\end{equation}
and the contact tangential force is 
\begin{equation}
	\mathbf{f}_{ti} = \mathbf{f}_{ci} -\mathbf{f}_{Ni}.
	\label{eq:ft}
\end{equation}

\subsection{Problem Description}
We define $\pfAll = [\mathbf{p}_{f1}^\myT, \mathbf{p}_{f2}^\myT, ... , \pf^\myT]^\myT$ to be the stacked vector of all the fingertip positions, and similarly $\pBfAll$ to be all the fingertip positions relative to the object and $\paAll$ to be all the finger anchor positions.  The duration of the regrasp is $T$.  

\textbf{Given: }the initial grasp configuration $\{\To(0)$, $\pfAll(0)$, $\paAll(0)\}$, the finger stiffness properties, the geometry of the rigid object and stationary environment, and the goal fingertip relative positions $\mathbf{p}^\mathcal{B}_{f, \,\text{goal}}$,

\textbf{Find: } motions of the object $\To(t)$ and finger anchors $\paAll(t)$ such that $\pBfAll(T) = \mathbf{p}^\mathcal{B}_{f, \,\text{goal}}$ and the rigid-body conditions and quasistatic force-balance conditions are satisfied at all times, $0\leq t \leq T$.

If the task involves carrying the object away from the rigid environment after the regrasp, the goal fingertip and anchor positions $\mathbf{p}^\mathcal{B}_{f, \,\text{goal}}$ and $\mathbf{p}_{a,\text{goal}}$ should be chosen to achieve force closure on the object, or at least to balance the object's gravitational wrench, without the benefit of the environmental contacts.  Note also that the robot itself can provide all or a portion of the stationary, rigid environment, e.g., using its palm or another link of the robot arm.

Because we assume quasistatic mechanics, the time variable $t$ in the problem formulation can be rescaled without affecting the spring-sliding regrasp.

\section{Finger Spring Compliance Model}
\label{sec:finger_compliance_model}

\begin{figure}[t]
	\centering
	\includegraphics[width = 3.4in]{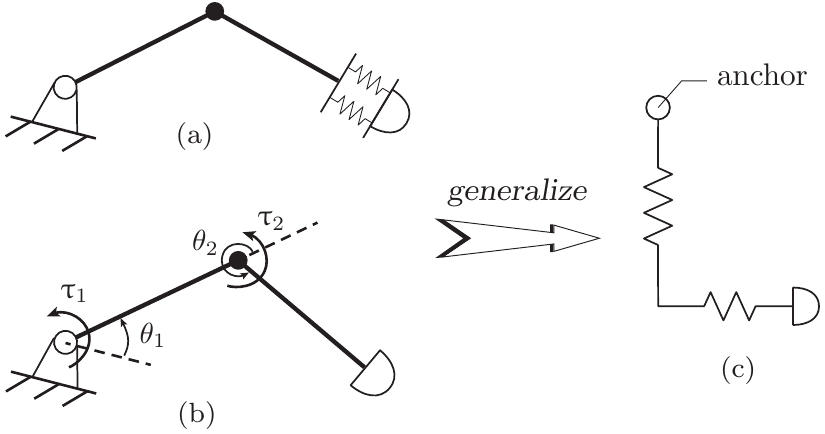}
	\caption{Generalized finger spring compliance model. (a) A compliant fingertip at the endof an otherwise rigid position-controlled finger. (b) A finger where the effective stiffness comes from active stiffness control, compliance at the joints or links, or joint-torque control and the geometry of the finger. Fingertip stiffnesses of both cases can be modeled by (c).}
	\label{fig:finger_springmodel}
\end{figure}

The springy-finger model can represent several different mechanical finger designs and control strategies.  For example, Figure~\ref{fig:finger_springmodel} shows two different types of fingers. In Figure~\ref{fig:finger_springmodel}(a), there is a spring-mounted fingertip attached to the end of a position-controlled finger (e.g., a stiff, highly geared finger). The anchor point is at the attachment of the spring to the finger.  This design directly matches our model provided the 3D stiffness of the spring is known.

Figure~\ref{fig:finger_springmodel}(b) represents the case where the fingertip is rigidly mounted to the finger. The effective stiffness may come from an active stiffness control law or from passive compliance at the joints (as with series elastic actuators) or at the links. Another interesting case 
occurs when passive compliance derives from open-loop torque-controlled joints of the finger.  In this case, the anchor is the base of the finger and the entire finger acts as a nonlinear spring.  Under certain circumstances, the linearized passive compliance at the contact is positive definite, as required by the assumptions.  This case is examined in more detail in Appendix~\ref{app:torque}.

\section{Finger Contact Mechanics}
\label{sec:contact_mechanics}%

This section answers the following question: given the object's motion and the $i$th finger anchor and contact positions, what is the relationship between the finger anchor velocity $\Dpa$ and the corresponding fingertip velocity $\Dpf$? 

Given the anchor and contact locations, the contact force is determined by the spring compliance.
The fingertip sticks to the object when (1) the contact force is in the interior of the friction cone or (2) the contact force is on the boundary of the friction cone but the anchor velocity results in a rate of change of the contact force that keeps it within the friction cone under the assumption of a stationary contact. 
If these conditions do not hold, the fingertip contact force is on the boundary of the friction cone and the tangential sliding velocity is aligned with the tangential contact force. 
For the sliding case, the \emph{forward mechanics} problem is to find the contact point velocity $\Dpf$ given the anchor velocity $\Dpa$, and the \emph{inverse mechanics} problem is to find the set of anchor velocities $\Dpa$ corresponding to a contact point velocity $\Dpf$.  Forward mechanics is useful for simulation, and inverse mechanics is useful for motion planning.

\begin{figure*}
	\centering
	\includegraphics[width = 7.in]{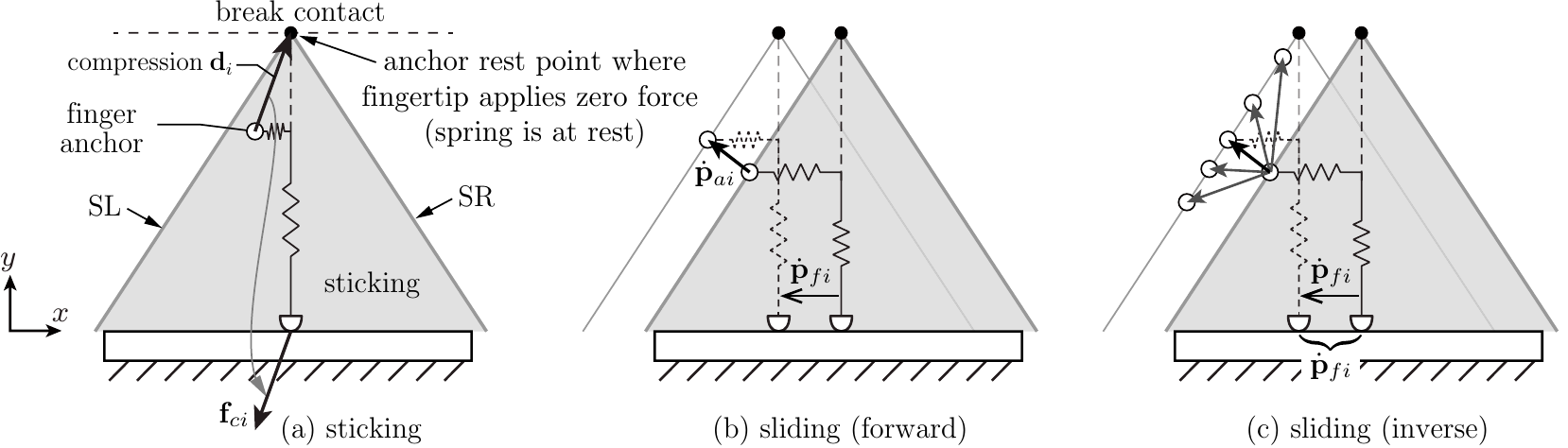}
	\vspace{-0.in}
        \caption{Illustration of planar forward and inverse contact mechanics for a finger in contact with a stationary object.  (a) The finger anchor is connected to the fingertip by a two-dimensional spring.  The spring compression is $\mathbf{d}_i$ and the contact force $\mathbf{f}_{ci}$ is determined by the stiffness matrix and $\mathbf{d}_i$.  If the anchor is anywhere inside the gray cone, the finger remains sticking to the object.  (This gray cone is just the friction cone at the fingertip contact, translated to the anchor rest point, when the spring stiffness matrix is a scalar multiple of the $2 \times 2$ identity matrix, $\K = k \mathbf{I}$.)  If the anchor goes above the dotted line, the contact normal force goes to zero and the fingertip breaks contact with the object.  If the anchor moves continuously to the left, when it reaches the left cone edge marked SL, the fingertip begins to slide to the left, and the cone moves along with the anchor.  A symmetric situation happens if the anchor moves right.  (b) Forward sliding mechanics:  If the anchor is at the left edge of the cone and moves with the velocity $\dot{\mathbf{p}}_{ai}$, then the fingertip slides with velocity $\dot{\mathbf{p}}_{fi}$.  (c) Inverse sliding mechanics:  Any of the anchor velocities indicated results in the leftward fingertip velocity $\dot{\mathbf{p}}_{fi}$ shown.  Note that some of the anchor velocities even have a component to the right, because these anchor velocities simultaneously reduce the normal force.}

\label{fig:1fingerEx}
\end{figure*}

The contact mechanics problems are illustrated by a simple example in Figure~\ref{fig:1fingerEx}. 

\subsection{Sticking Case} 

When fingertip $i$ sticks to the object, the fingertip follows the object's motion, i.e.,
\begin{equation}
	\dot{\mathbf{p}}_{fi}^{\mathcal{B}} = \mathbf{0}.
	\label{eq:relative_vel=0}
\end{equation}
The transformations of the contact position and velocity from $\mathcal{B}$ to $\mathcal{W}$ can be written as
\begin{align}
	\mathbf{p}_{fi} &= \mathbf{p}_o + \mathbf{R}_o\mathbf{p}_{fi}^{\mathcal{B}} \;, \\
	\dot{\mathbf{p}}_{fi} &= \dot{\mathbf{p}}_o + \dot{\mathbf{R}}_o\mathbf{p}_{fi}^{\mathcal{B}} +\mathbf{R}_o \dot{\mathbf{p}}_{fi}^{\mathcal{B}} .
	\label{eq:finger_vel_trans}
\end{align}
Substituting Equation~\eqref{eq:relative_vel=0} into Equation~\eqref{eq:finger_vel_trans}, the fingertip velocity in $\mathcal{W}$ is 
\begin{equation}
	\dot{\mathbf{p}}_{fi} = \dot{\mathbf{p}}_o + \dot{\mathbf{R}}_o\mathbf{p}_{fi}^{\mathcal{B}} = \dot{\mathbf{p}}_o + \bm\omega_o \times \mathbf{R}_o\mathbf{p}_{fi}^{\mathcal{B}}.
	\label{eq:sticking_finger_vel}
\end{equation}

\subsection{Sliding Case}
\subsubsection{Forward Mechanics}
\label{subsec:contact_mechanics_forward}

When sliding, the contact forces of the $i$th finger satisfy
\begin{equation}
	\|\mathbf{f}_{ti}\| = \mu \|\mathbf{f}_{Ni}\|.
	\label{eq:sliding_condition}
\end{equation}
We define the finger sliding velocity relative to $\mathcal{B}$ as 
\begin{equation}
	\dot{\mathbf{p}}_{fi}^{\mathcal{B}} = \lambda_i \mathbf{f}_{ti}^{\mathcal{B}} = \lambda_i \mathbf{R}_o^\myT \mathbf{f}_{ti},
	\label{eq:sliding_vel_relation}
\end{equation}
which enforces the Coulomb friction assumption that the sliding velocity is in the direction of the tangential frictional force applied by the finger to the object.  The positive scalar $\lambda_i$, which must be solved for, relates 
the magnitudes of the friction force and the sliding velocity.

Substituting Equation~\eqref{eq:sliding_vel_relation} into \eqref{eq:finger_vel_trans}, we have
\begin{align}
	\dot{\mathbf{p}}_{fi} =& ~\Dpo + 
	\dot{\mathbf{R}}_o \mathbf{p}_{fi}^{\mathcal{B}} + \mathbf{R}_o \lambda_i \mathbf{R}_o^\myT \mathbf{f}_{ti} \nonumber\\
	 =& ~\mathbf{c}_{fi} + \lambda_i \mathbf{f}_{ti},
	\label{eq:finger_vel}
\end{align}
where $\mathbf{c}_{fi} = \Dpo + [\bm \omega_o] \mathbf{R}_o  \mathbf{p}_{fi}^{\mathcal{B}}$ reflects the change of the contact point position due to the object motion, without sliding.

From Equation~\eqref{eq:sliding_condition}, we find
\begin{align}
	\|\mathbf{f}_{ti}\| \|\mathbf{f}_{ti}\| &= \mu^2 \|\mathbf{f}_{Ni}\| \|\mathbf{f}_{Ni}\| \nonumber \\
	\rightarrow \mathbf{f}_{ti} \cdot \mathbf{f}_{ti} &= \mu^2 \: \mathbf{f}_{Ni} \cdot \mathbf{f}_{Ni} \nonumber \\
	 \xrightarrow[]{\frac{d}{dt}} \dot{\mathbf{f}}_{ti} \cdot \mathbf{f}_{ti} + \mathbf{f}_{ti} \cdot \dot{\mathbf{f}}_{ti} &= \mu^2( \dot{\mathbf{f}}_{Ni} \cdot \mathbf{f}_{Ni} + \mathbf{f}_{Ni} \cdot \dot{\mathbf{f}}_{Ni}) \nonumber \\
	 \rightarrow \mathbf{f}^\myT_{ti} \dot{\mathbf{f}}_{ti} &= \mu^2 \:  \mathbf{f}^\myT_{Ni} \dot{\mathbf{f}}_{Ni}.
	 \label{eq:sliding_condition_d}
\end{align}
Then 
from Equation~\eqref{eq:n^hat} we have
\begin{equation}
	\Dnhat = \frac{\partial \nhat}{\partial \pBf }\DpBf = \frac{\partial \nhat}{\partial \pBf } \lambda_i \fBt = \lambda_i \gn,
	\label{eq:Dnhat}
\end{equation}
where $\gn = \frac{\partial \nhat}{\partial \pBf } \Ro^\myT \ft$ and $\frac{\partial \nhat}{\partial \pBf }$ represents the curvature of the object at the contact point.

In some cases, such as a linear-spring-mounted fingertip as in Figure~\ref{fig:finger_springmodel}(a), the finger's stiffness matrix $\K$ is constant.  In general, the stiffness matrix may be a function of the finger contact location $\pf$ and other parameters $\bm\sigma$  used to control the stiffness (as in variable-stiffness actuators).
In this case, the stiffness can be written $\K(\pf, \bm\sigma)$, and taking the derivative of Equation~\eqref{eq:fc} and combining with Equation~\eqref{eq:finger_vel} gives
\begin{align}
	\Dfc &= -\DK \di - \K(\Dpf - \Dpa) \nonumber \\
	 &= -\left(\frac{\partial \K}{\partial \pf} \Dpf + \frac{\partial \K}{\partial \bm\sigma} \dot{\bm\sigma} \right) \di - \K (\cf + \lambda_i \ft) + \K \Dpa \nonumber \\
	 &= \lambda_i \gc + \cc,
	 \label{eq:Dfc}
\end{align}
where $ \gc = -\K \ft - \frac{\partial \K}{\partial \pf} \ft \di$, and $\cc = \K\Dpa - ( \frac{\partial \K}{\partial \bm\sigma}\dot{\bm\sigma} + \frac{\partial \K}{\partial \pf} \cf ) \di - \K \cf$. By denoting $\hi = \left(\frac{\partial \K}{\partial \bm\sigma}\dot{\bm\sigma} + \frac{\partial \K}{\partial \pf} \cf \right)\di + \K \cf$, we have $\cc = \K\Dpa - \hi$.
In the case that $\K$ is constant, Equation~\eqref{eq:Dfc} simplifies to
\begin{equation}
  \Dfc = \K (\Dpa - \Dpf) = -\K \dot{\mathbf{d}}_i.
  \end{equation}

Taking the derivative of Equations~\eqref{eq:fN} and \eqref{eq:ft} and combining with Equations~\eqref{eq:Dnhat} and \eqref{eq:Dfc} yields
\begin{align}
	\DfN &= \Dfc^\myT \nhat \nhat + \fc^\myT \Dnhat \nhat +  \fc^\myT \nhat \Dnhat \nonumber \\
	&= (\lambda_i \gc + \cc)^\myT \nhat \nhat + \fc^\myT \lambda_i \gn \nhat + \fc^\myT \nhat \lambda_i \gn \nonumber \\
	&= \lambda_i \gN + \cN
	\label{eq:f_Ni_dot} 
\end{align}
where $\gN = \gc^\myT \nhat \nhat + \fc^\myT \gn \nhat + \fc^\myT \nhat \gn$, $\cN = \cc^\myT \nhat \nhat$, and
\begin{align}
	\Dft = \Dfc - \DfN = \lambda_i\gt + \ct,
	\label{eq:f_ti_dot}
\end{align}
where $\gt = \gc - \gN$ and $\ct = \cc - \cN$.

Substituting Equations~\eqref{eq:f_Ni_dot} and \eqref{eq:f_ti_dot} into \eqref{eq:sliding_condition_d} we can solve for $\lambda_i$ as
\begin{equation}
	\lambda_i = \frac{\mu^2 \fN^\myT \cN - \ft^\myT \ct}{\ft^\myT \gt - \mu^2 \fN^\myT \gN}.
	\label{eq:lambda1}
\end{equation}

In the numerator, since $\cN = \cc^\myT \nhat \nhat = (\cc \cdot \nhat) \nhat$ is along the contact normal, the term $\fN^\myT \cN$ is equivalent to $\fN^\myT \cc$. 
By plugging in $\ct = \cc - \cN$, Equation~\eqref{eq:lambda1} simplifies to
\begin{align}
	\lambda_i &= \frac{\mu^2 \fN^\myT \cc - \ft^\myT \cc + \cancelto{0\text{ (orthogonal)}}{\ft^\myT \cN} }{\ft^\myT \gt - \mu^2 \fN^\myT \gN} 
	 = \frac{\ai^\myT \cc }{\lden} \nonumber \\
	 &= \frac{\ai^\myT \K}{\lden}\Dpa - \frac{\ai^\myT \hi}{\lden} \nonumber \\
	 &= \glan \Dpa - \clan,
	 \label{eq:lambda2}
\end{align}
where $\ai = \mu^2 \fN - \ft \in \real^{3 \times 1}$ , $\lambda_{\text{den},i} = \ft^\myT \gt - \mu^2 \fN^\myT \gN$, $\glan = \ai^\myT \K /\lden \in \real^{1 \times 3}$, and $\clan = \ai^\myT \hi /\lden$.
The finger contact sliding velocity $\Dpf$ can be solved for by substituting $\lambda_i$ into Equation~\eqref{eq:finger_vel}.

\subsubsection{Inverse Mechanics}
\label{subsec:contact_mechanics_inverse}

\begin{figure}
	\centering
	\includegraphics[width=2.5in]{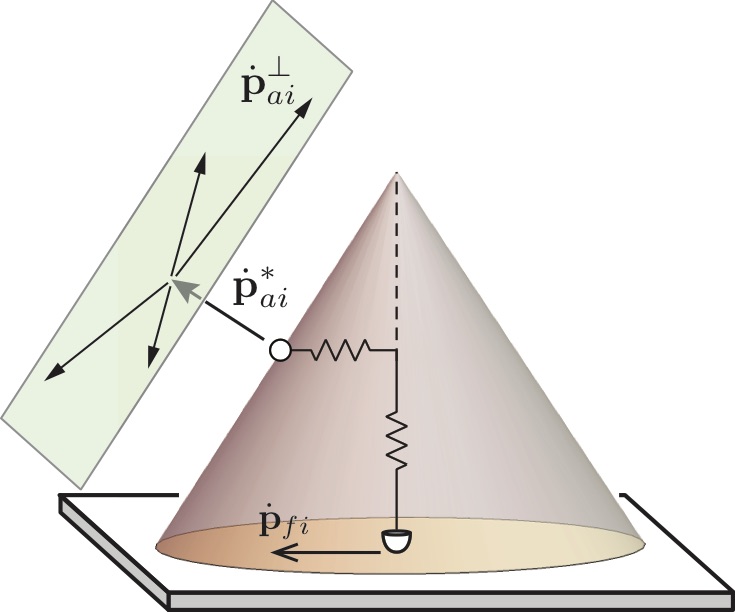}
	\vspace{-0.in}
        \caption{A 3D version of Figure~\ref{fig:1fingerEx}(c).  The direction of fingertip sliding $\Dpf$ is determined by the current force on the boundary of the friction cone, and the magnitude $\|\Dpf\|$ of the desired sliding velocity places one constraint on the anchor velocity, resulting in a plane of anchor velocities $\Dpa$ that achieve the desired fingertip sliding velocity $\Dpf$.  This plane is defined by the sum of a particular solution $\Dpa^*$ and any $\Dpa^\perp$ in the two-dimensional space spanned by $\mathbf{I} - \glanInv \glan$ (Equation~\eqref{eq:Dpa}).}
        \label{fig:1fingerEx3d}
\end{figure}

The result of the forward mechanics, Equation~\eqref{eq:lambda2}, gives the finger sliding 
velocity for a given finger anchor velocity. For the inverse mechanics problem, we solve for the anchor motions $\Dpa$ that cause a desired finger contact sliding velocity $\Dpf$. 
Since the object motion and the contact force are known, a desired finger contact sliding velocity $\Dpf$ is equivalent to a desired $\lambda_i$ from Equation~\eqref{eq:finger_vel}. 
Therefore we can write all solutions to the inverse problem as
\begin{align}
  & \Dpa =  \Dpa^{*} + \Dpa^\perp, \label{eq:Dpa} \\
  \text{where } & \Dpa^{*} = \glanInv (\lambda_i + \clan) \text{ and } \nonumber \\
  & \Dpa^\perp \in \{(\mathbf{I}^{3\times3} - \glanInv \glan) \mathbf{v} \; | \;  \mathbf{v} \in \real^3\}. \nonumber
\end{align}
The vector $\Dpa^*$ is a particular solution for $\Dpa$ found using the pseudoinverse  $\glanInv = \glanT (\glan \glanT)^{-1} = \glanT / \|\glan\|^2$ and $\Dpa^\perp$ is any vector in the two-dimensional space spanned by $\mathbf{I} - \glanInv \glan$, the space of anchor velocities that have no impact on the fingertip sliding velocity.  Figure~\ref{fig:1fingerEx3d} illustrates the space of anchor velocity solutions for a 3D version of Figure~\ref{fig:1fingerEx}(c).

For different solutions of $\Dpa$, all the corresponding contact sliding velocities are the same but the changes of the contact force $\Dfc$ are different.
By Equation~\eqref{eq:Dfc} we can solve the corresponding contact force change $\Dfc$ for each $\Dpa$. 
The redundancy resolution in the choice of $\Dpa$ could be based on additional constraints on the anchor motions or optimization of desired contact force properties.

\subsubsection{Degenerate Cases}
\label{subsec:sliding_ill_condition}

\begin{figure}
	\centering
	\includegraphics{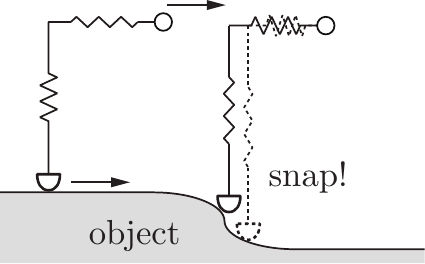}
	\vspace{-0.in}
        \caption{A springy finger dragged over a rounded ledge may suddenly slide dynamically before quasistatic motion resumes.}
        \label{fig:degeneracy}
\end{figure}

In quasistatic sliding, Equations \eqref{eq:lambda2} (coupled with Equation~\eqref{eq:finger_vel}) and \eqref{eq:Dpa} describe the relationship between the anchor motion and the contact point motion. Two degeneracies are possible, when (I) $\glan = 0$ or (II) $\lden = 0$.  For a degeneracy of type I, the anchor velocity has no impact on the sliding velocity of the fingertip.  For a degeneracy of type II, the fingertip velocity becomes unbounded and the quasistatic assumption is violated.  An example of a degeneracy of type II is shown in Figure~\ref{fig:degeneracy}.

As shown in Proposition~\ref{prop:nonzero_glan}, a degeneracy of type I cannot occur under our assumptions.

\begin{proposition}
The 3-vector $\glan$ is nonzero under Assumptions \ref{enum:stiff_assumption}) and \ref{enum:fc_assumption}) (the finger has a positive-definite stiffness matrix $\K$ and maintains a positive contact normal force). 
\label{prop:nonzero_glan}
\end{proposition}
\begin{IEEEproof}
The vector $\glan$ is proportional to the product $\ai^\myT \K$.  By Assumption~\ref{enum:fc_assumption}) the term $\ai = \mu^2 \fN - \ft$ will be nonzero.  Proposition~\ref{prop:nonzero_glan} holds since the matrix $\K$ is full rank when it is positive definite.
\end{IEEEproof}

\vspace{0.05in}

Considering degeneracies of type II, many factors affect the value of $\lden$, including the local curvature of the object and variations in the finger stiffness.  In the particular case that the stiffness $\K$ is constant and the object surface is flat, however, this type of degeneracy cannot occur under our assumptions.

\begin{proposition}
	When the finger stiffness matrix $\K$ is constant and the local curvature of the object at the contact is zero, $\lden$ will be nonzero under Assumptions \ref{enum:stiff_assumption}) and \ref{enum:fc_assumption}) (the finger has a positive-definite stiffness matrix $\K$ and maintains a positive contact normal force). 
	\label{prop:nonzero_lambda}
\end{proposition}

\begin{IEEEproof}
	When $\frac{\partial \K}{\partial \pf}=0$, $\frac{\partial \K}{\partial \bm\sigma}=0$ and $\frac{\partial \nhat}{\partial \pBf }=0$, the key variables in Equations~\eqref{eq:f_Ni_dot} and \eqref{eq:f_ti_dot} are
	\begin{equation}
		\gc = -\K \ft \text{ and } ~\gN = -\ft^\myT \K^\myT \nhat \nhat .
		\label{eq:gcN_simp}
	\end{equation}
	Because $\gN$ and $\fN$ are both vectors in the direction of $\nhat$, we have $\fN^\myT \gN = -\|\fN\| \ft^\myT \K^\myT \nhat$.
	Plugging Equation~\eqref{eq:gcN_simp} into \eqref{eq:lambda2}, we have	
	\begin{align*}
		\lden &= \ft^\myT (\gc-\gN) - \mu^2 \fN^\myT \gN \\
			  			  &= -\ft^\myT \K \ft + \mu^2 \ft^\myT \K^\myT \fN.
	\end{align*}
	Since $\K$ is symmetric,
	\begin{align*}
		\lden &=  \ft^\myT \K (\mu^2\fN-\fthat) = \ft^\myT \K \ai, \numberthis
	\end{align*}
	where $\ft$ and $\ai$ are both nonzero due to Assumption~\ref{enum:fc_assumption}). Similar to the proof of Proposition~\ref{prop:nonzero_glan}, since $\K$ is positive definite, $\lden$ is nonzero.
\end{IEEEproof}

\section{Object Mechanics}
\label{sec:obj_mechanics}

The grasped object has $m$ point contacts with the rigid stationary environment, and according to the planned object motion $\mathbf{T}_o(t)$, $t \in [0,T]$,
each contact could be sliding (relative motion at the point of contact) or rolling/sticking (no sliding at the contact).  At each sliding contact, the total contact force applied to the object lies on a one-dimensional line on the boundary of the friction cone, such that the tangential frictional force is opposite the direction that the object slides relative to the environment and has a magnitude $\|\mu f_N\|$ (where $f_N$ is the normal force).  At each sticking or rolling contact,
the contact force lies somewhere inside the three-dimensional circular friction cone.
In other words, a sliding contact offers one force freedom and a sticking contact offers three force freedoms to satisfy quasistatic wrench balance, which requires the wrenches from the environmental contacts, the finger contacts, and gravity to sum to zero.

If the $j$th external contact with the environment is sticking or rolling, the friction cone can be approximated as an $n_c$-sided polyhedral cone, i.e., the nonnegative linear combination of $n_c$ unit forces on the boundary of the circular friction cone, $\hat{\mathbf{f}}_{jk}, k = 1, \ldots, n_c$.
Given the contact location $\pe$ expressed in $\mathcal{W}$, each of these forces corresponds to a wrench $\wkhat = [(\pe \times \fkhat)^\myT, \fkhat^\myT]^\myT \in \real^6$, and the nonnegative linear combination of the $n_c$ wrenches is the wrench cone $\mathcal{WC}_{ej}$.  The contact wrench at the $j$th contact point can be expressed as
\begin{equation}
	\we = \sum_{k=1}^{n_c} \beta_{jk} \wkhat,\; \beta_{jk} \geq 0,
\end{equation}
where the nonnegative $\beta_{jk}$ coefficients multiply the wrench cone edges to yield the total contact wrench (see, e.g.,~\cite{Kao2016,Lynchbook2017}).

If the $j$th contact is sliding, it provides a single unit force $\hat{\mathbf{f}}_{j1}$ on the friction cone, which corresponds to a single contact wrench $\mathbf{w}_{j1}$ and a single free coefficient $\beta_{j1} \geq 0$ multiplying it, i.e., $\mathbf{w}_{ej} = \beta_{j1} \mathbf{w}_{j1}$.

We denote $\weAll$ as the sum of all the external contact wrenches, 
\begin{equation}
	\weAll = \sum_{j=1}^m \we  = \What \bm \beta \in \mathcal{WC}_e,  
	\label{eq:weALL}
\end{equation} 
where $\mathcal{WC}_e$ is the wrench cone for all external contacts, $\What \in \mathbb{R}^{6 \times p}$ consists of the $p$ column vectors of the individual contact wrench cone edges, and $\bm\beta \in \mathbb{R}^{p \times 1}$ is a column vector of the corresponding nonnegative wrench coefficients.

For the finger contact force $\fc$, the corresponding wrench applied to the object is 
\begin{equation}
	\wc = [(\pf \times \fc)^\myT, \fc^\myT]^\myT.
	\label{eq:total_wc}
\end{equation}
The object wrench-balance condition can be written as
\begin{equation}
	\wcAll + \weAll + \wg = \mathbf{0},
	\label{eq:ext_force_bal}	
\end{equation}
where $\wcAll = \sum_{i=1}^n \wc$ is the total finger contact wrench and $\wg$ is the gravitational wrench.

\section{Robustness Analysis}
\label{sec:robustness}

At a given time during execution of a planned sliding regrasp, the expected finger contact wrench on the object is $\wcbar$, but due to uncertainty in friction, anchor motions, and contact geometry, the actual contact wrench is assumed to be $\wcAll = \wcbar + \dwc$, where $\dwc$ is a disturbance.

\begin{definition}
  A planned regrasp is \emph{robust to $\varepsilon$ wrench uncertainty} (or \emph{$\varepsilon$-robust} for short) if, for all $t\in[0,T]$, there exists a $\weAll(t) \in \mathcal{WC}_e(t)$ such that
  \begin{equation}
    \bar{\mathbf{w}}_c(t) + \dwc(t) + \weAll(t) + \wg(t) = \mathbf{0},
    \label{eq:uncertain_balance}
  \end{equation}
  where $\wcbar(t)$ is the expected fingertip wrench during the regrasp and each of the six components of the fingertip wrench disturbance $\dwc(t)$ can take any value in the range $[-\varepsilon,\varepsilon], \varepsilon >0$.
	\label{def:robust_definition}
      \end{definition}

\begin{remark}
  The definition of $\varepsilon$-robustness does not differentiate between forces and moments in a wrench.  Moments can be divided by a characteristic length-scale factor to have the same units as forces. 
\end{remark}

\begin{remark}
Since $\varepsilon$-robustness is based on full-dimensional wrench uncertainty at the fingertips, it also implies robustness to small wrench uncertainty at the environmental contacts.
\end{remark}

\begin{figure}
	\centering
	\includegraphics[width = 2.8in]{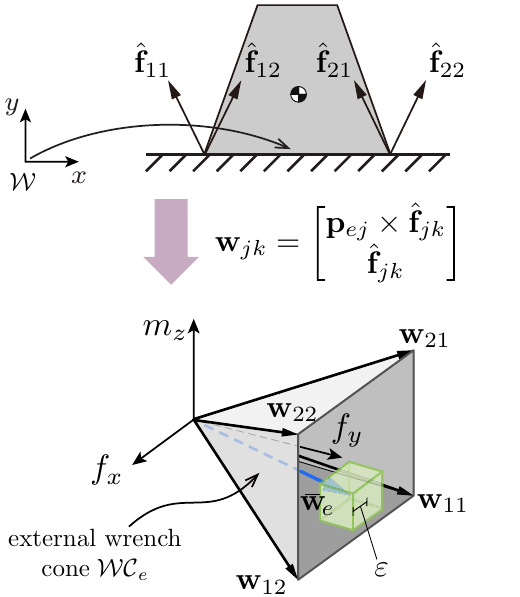}
	\caption{
 A 2D example showing robustness to $\varepsilon$ wrench uncertainty. An object sits on a table with a line contact. The four basis wrench vectors form the external wrench cone $\WCe$. 
          The blue arrow shows the nominal wrench-balancing external wrench $\bar{\mathbf{w}}_e$, at the center of the green $2\varepsilon \times 2 \varepsilon \times 2 \varepsilon$ cube of uncertain external wrenches $\dwe$ satisfying quasistatic wrench balance, i.e., $\dwe = -\dwc$. 
          The uncertainty cube is fully contained in $\WCe$, so the plan is
          robust to $\varepsilon$ wrench uncertainty at this instant.
        }
	\label{fig:2D_wrench_cone}
\end{figure}

A planar example is shown in Figure~\ref{fig:2D_wrench_cone}. The four external basis wrench vectors give the external wrench cone $\WCe$. For the nominally-required external wrench $\bar{\mathbf{w}}_e$, as long as the $\varepsilon$ wrench uncertainty cube is within the external wrench cone $\WCe$, the plan is robust to $\varepsilon$ wrench uncertainty at this instant.

A necessary and sufficient condition for $\varepsilon$-robustness is that each of the $2^6$ corners of the wrench disturbance hypercube lies within $\WCe$.  Proposition~\ref{prop1} gives a simple sufficient condition for $\varepsilon$-robustness.

\begin{proposition}
  A planned regrasp is $\varepsilon$-robust if $\What(t)$ has rank six and the planned nominal fingertip wrench $\wcbar(t)$ permits a nominal environmental wrench coefficient vector $\bar{\bm \beta}(t)$ satisfying
  \vspace*{0.1in}
  
  \begin{tabular}{rrl}
    \hspace*{-0.2in} nominal wrench balance: \hspace*{-0.1in} & $\wcbar(t) + \What(t) \bar{\bm \beta}(t) + \wg(t)$ \hspace*{-0.15in} & $= \mathbf{0}$ \\
    \hspace*{-0.2in} robustness: \hspace*{-0.1in} & $\bar{\bm \beta}(t) - \varepsilon \|\What^\dagger(t)\| \mathbf{1}$ \hspace*{-0.15in} & $\geq \mathbf{0}$,
    \end{tabular}
    \vspace*{0.1in}
    
    \noindent
    for all $t \in [0,T]$, where $\What^\dagger(t) = \WhatT(t) (\What(t) \WhatT(t))^{-1}$, $\mathbf{0}$ and $\mathbf{1}$ are vectors of zeros and ones, and $\|\cdot \|$ is the matrix norm induced by the vector 2-norm.
  \label{prop1}
\end{proposition}

\begin{IEEEproof}
  Since the uncertainty $\dwc(t)$ spans all dimensions of the wrench space, the rank of $\What(t)$ must be six.
	
	From Definition~\ref{def:robust_definition} and Equation~\eqref{eq:weALL}, at any given time wrench balance with uncertainty requires 
\begin{equation*}
  \What \dbeta  = -\dwc,
\end{equation*}
where $\bar{\bm \beta} + \dbeta = \bm \beta$ defines an environmental contact wrench $\What \bm \beta$ satisfying wrench balance when including the disturbance $\dwc$.
A particular solution to this equation is
\begin{equation}
  \dbeta = -\What^\dagger \dwc.
  \label{eq:dbeta}
\end{equation}
To satisfy the Coulomb friction assumption, we have
\begin{equation}
	\bar{\bm \beta} + \dbeta \geq \mathbf{0}.
	\label{eq:friction_condi}
\end{equation}
Substituting Equation~\eqref{eq:dbeta} into \eqref{eq:friction_condi} gives
\begin{equation}
	\bar{\bm \beta} - \What^\dagger \dwc \geq \mathbf{0}.
	\label{eq:beta_condi}
      \end{equation}
      Since each component of $\dwc$ must be in the range $[-\varepsilon,\varepsilon]$,
      \begin{equation}
        \varepsilon \|\What^\dagger\| \mathbf{1} \geq \What^\dagger \dwc
        \label{eq:ineq}
      \end{equation}
      and the robustness condition in the proposition follows by substituting~\eqref{eq:ineq} into \eqref{eq:beta_condi}.
    \end{IEEEproof}

The robustness condition in Proposition~\ref{prop1} implies that $\varepsilon$-robustness can be obtained for larger values of $\varepsilon$ if the environmental contact wrench coefficients $\bar{\bm \beta}$ are larger. Since larger environmental wrenches imply larger fingertip wrenches by quasistatic wrench balance, fingers with greater force-generation capability are generally capable of larger values of $\varepsilon$-robustness.

$\varepsilon$-robustness requires a full-dimensional external wrench cone $\WCe$, i.e., a wrench cone with a non-empty interior.  This can be achieved by two frictional rolling/sticking contacts in the plane or three frictional rolling/sticking contacts in 3D.  While it is possible to have a full-dimensional external wrench cone when one or more contacts roll or slide, such cases are exceptions, relying on very specific contact geometries.  For this reason, in the remainder of the paper we focus on the case where $\varepsilon$-robustness is achieved by the object remaining stationary relative to the rigid environment.

\section{Sliding Regrasp Planning}
\label{sec:motion_planning}

The finger and object mechanics of the previous sections provide constraints that must be satisfied by a sliding regrasp plan.  A planning algorithm may be expressed
as a constraint satisfaction problem or
as a constrained optimization, as in Table~\ref{table:generic}.

\begin{table}
  \begin{center}
  \normalsize
  \frame{
  \begin{tabular}{rl}
  {\bf find} & $\mathbf{T}_o(t), \pfAll(t), \paAll(t)$, and $T$, $t \in [0,T]$ \\
  {\bf maximizing} & $\varepsilon$ robustness (Section~\ref{sec:robustness}) \\
  {\bf such that} & 1) $\{\mathbf{T}_o(0), \pfAll(0), \paAll(0)\}$ is the original \\
  & \hspace*{0.2in} grasp configuration \\
  & 2) $\mathbf{T}_o(T)$, $\pfAll(T)$ is the desired grasp \\
  &  3) finger contact mechanics are satisfied \\
  & \hspace*{0.2in} for all $t \in [0,T]$ (Section~\ref{sec:contact_mechanics}) \\
  & 4) object wrench balance is achieved \\
  & \hspace*{0.2in} for all $t \in [0,T]$ (Section~\ref{sec:obj_mechanics}) \\
  & 5) fingertip and anchor kinematic \\
  & \hspace*{0.2in}  constraints, force/velocity bounds, \\
  & \hspace*{0.2in} and other constraints are satisfied \\
  & 6) (optional) $\{\mathbf{T}_o(T), \pfAll(T), \paAll(T)\}$ is \\
  & \hspace*{0.2in} a force-closure grasp without external\\
  & \hspace*{0.2in} contacts, for a subsequent carry
  \end{tabular}
  }
\end{center}
  \caption{An example sliding regrasp planning formulation.}
  \label{table:generic}
\end{table}

The wrench-balance constraint 4) is redundant with the optimization criterion:  if the maximum $\varepsilon$ to which the plan is robust is greater than zero, then constraint 4) is automatically satisfied.  The regrasp planning problem could be reformulated to encode a robustness condition in constraint 4) and to change the objective function to minimize forces applied by the fingertips, as a way of resolving the finger contact inverse mechanics redundancy.  Or the objective function could be eliminated completely, turning the planning problem into a constraint satisfaction problem instead of an optimization.

How to efficiently implement the planner depends on properties of the
robot hand and other details that may be task-specific, and it is not the purpose of this paper to propose a single implementation for all tasks, objects, and robot hands.
Choices include how to represent the trajectories using finite parametrizations;
whether to use local gradient-based optimization methods based on collocation or shooting, global optimization methods, search-based methods; etc.  Instead of solving for both $\pfAll(t)$ and  $\paAll(t)$ and constraining them to be consistent, we could solve only for $\paAll(t)$ and use forward mechanics (Section~\ref{subsec:contact_mechanics_forward}) to determine the corresponding $\pfAll(t)$, or we could solve only for $\pfAll(t)$ and use inverse mechanics with redundancy resolution (Section~\ref{subsec:contact_mechanics_inverse}) to solve for $\paAll(t)$.  Also, in a typical regrasp plan, each fingertip starts out sticking while the anchor repositions itself to bring the contact force to the boundary of the friction cone; the fingertip transitions to sliding; and finally the fingertip reverts to sticking while the anchor is repositioned to bring the contact force to the interior of the friction cone, once the new grasp is achieved.  The planner can treat these segments (with their different finger contact mechanics) separately, subject to continuity constraints at the transitions.  Finally, we could restrict the object to be stationary ($\mathbf{T}_o(t) = \mathbf{T}_o(0)$) during the regrasp to achieve $\varepsilon$-robustness, following the discussion at the end of Section~\ref{sec:robustness}.

In the next section we describe one way to implement the general regrasp planning approach for the specific case of a two-fingered regrasp.

\section{Implementation}%
\label{sec:implementation}%

In this section we describe a two-fingered sliding regrasp task and an implementation of the regrasp planner of Table~\ref{table:generic}.  First we introduce the experimental setup; then we describe our methods for experimentally extracting relevant modeling parameters; and finally we give an implementation of the planning strategy outlined in Table~\ref{table:generic} as well as simulation and experimental results.

The experimental regrasp task was designed to be simple enough to yield insight into the derivations of the previous sections and to allow graphical interpretation of the robustness condition.
To satisfy $\varepsilon$-robustness, the fingers keep the object stationary during the regrasp.

\subsection{Experimental Regrasp Task}
\label{subsec:2Dsys_description}

\begin{figure}
  \centering
  \includegraphics[width=3in]{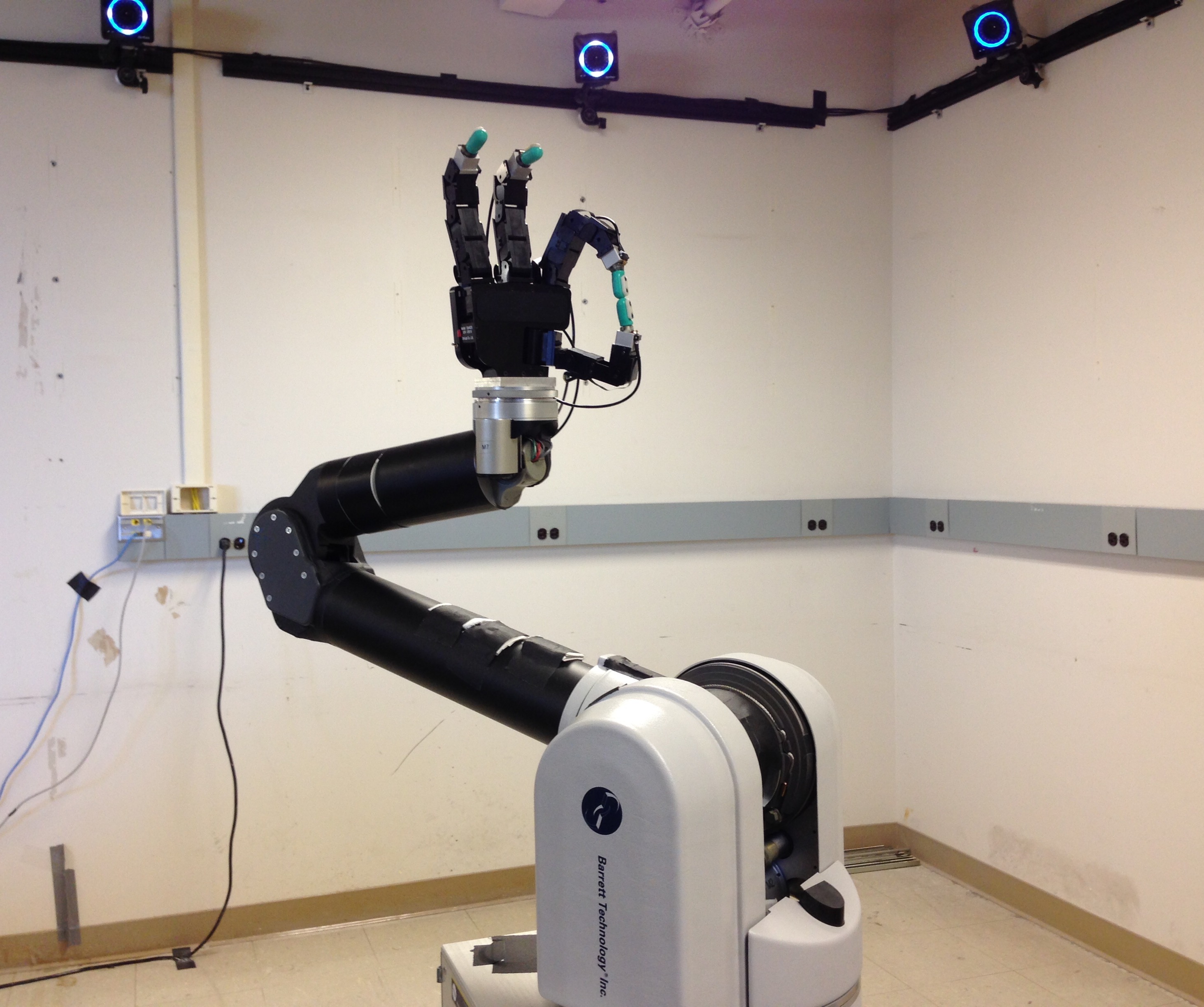}
  \caption{The ERIN manipulation system.}
  \label{fig:ERIN}
\end{figure}

\begin{figure}
	\centering
	\includegraphics[width = \colwidth]{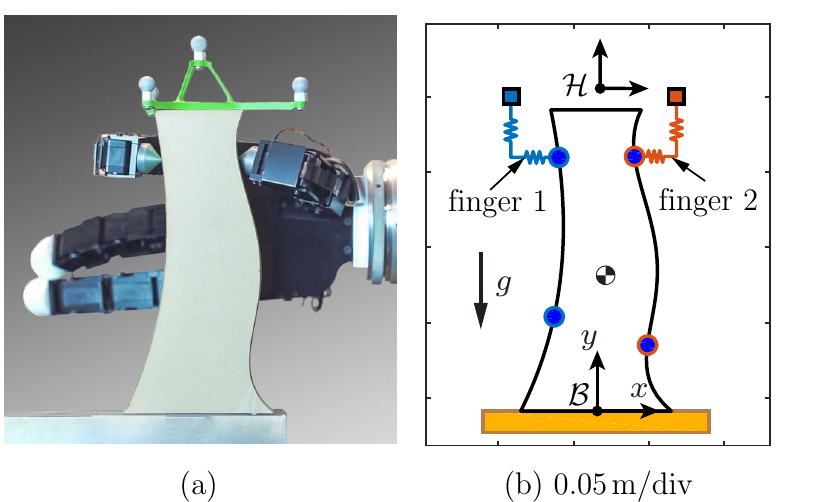}
	\vspace{-0.1in}
	\caption{Sliding regrasp task.  (a) The Allegro hand grasping an extruded object sitting on a table.  The hand is at its initial grasp configuration. (b) The positions of the virtual anchors (squares) are fixed in the hand frame $\mathcal{H}$. The colored lines between the anchors and fingertips (circles) show the programmed springs. The fingertips are shown at their initial configuration (top) and their desired regrasp configuration (bottom).  The object remains stationary during the sliding regrasp.}
	\label{fig:2Dconfig}
\end{figure}

For our experiments, we used our ERIN manipulation system, consisting of a ten-camera OptiTrack high-speed vision system, a Barrett WAM 7-dof arm, and a four-fingered Allegro robot hand with replaceable fingertips~\cite{Shi2017} (Figure~\ref{fig:ERIN}).  Two fingers of the hand grasp an object with smooth edges (Figure~\ref{fig:2Dconfig}). The object sits on a fixed table, and the motions of the hand are in the vertical plane. Figure~\ref{fig:2Dconfig} shows an initial configuration of the fingertips near the top of the object and a desired regrasp configuration near the bottom.  The friction coefficient between the object and the table is $\mu_e = 1$ and the gravitational force acting on the object is $10.1$\,N in the $-y$-direction.

Let $\mathcal{H}$ denote a frame attached to the hand with an origin at $\ph$. The finger stiffnesses and anchor positions are assumed to be fixed in $\mathcal{H}$, i.e., $\KH$ and $\pHa$ are constant. Therefore the anchor positions are uniquely determined by the hand configuration and in-hand sliding is realized by controlling the hand motion. For simplicity, we allow only $(x,y)$ translational hand motions in the vertical plane, so the anchor velocities are identical and confined to a two-dimensional space.  Under these constraints, the sliding inverse mechanics of Section~\ref{subsec:contact_mechanics_inverse} yields unique anchor velocities: the redundancies in the possible anchor velocities from the sliding inverse mechanics are resolved by the limited motions available to the hand (and therefore the finger anchors).  While the hand moves downward (in the $-y$-direction), the normal forces, relative sliding velocities at the two fingers, and $\varepsilon$-robustness can be modulated by the hand's motion in the $x$-direction.

The WAM arm controls the hand's motion at 500~Hz, and markers attached to the object and hand allow the vision system to track their 3D configurations at 360~Hz.

Each fingertip is a cone, yielding a well-defined contact point, and each finger consists of four joints individually controlled by geared DC motors.
The fingers are joint-torque controlled at $333$\,Hz to achieve the desired fingertip springiness $\KH$.  The constant virtual anchor location $\pHzero$ of finger $i$ relative to the hand is the controlled location of the fingertip when it applies zero force.  The rest length of the virtual spring is zero (i.e., $\mathbf{d}_{0i} = 0$), so the extension of the virtual spring is given by $\pHzero - \pHf$, where $\pHf$ is the actual fingertip location, and this spring extension is turned into finger reference joint torques by the equation
\begin{equation}
	\bm{\tau}_i = \mathbf{J}^\myT_i \left[ \KH (\pHzero - \pHf) \right],
\end{equation} 
where $\bm \tau_i$ 
denotes the joint torques for finger $i$ and $\mathbf{J}_i$ 
denotes the finger's Jacobian matrix.
Finger joint encoder feedback is used to evaluate $\pHf$ and $\mathbf{J}_i$.
\subsection{Parameter Identification}
\label{sec:param_iden}

\begin{figure}
	\centering
	\includegraphics[width = \colwidth]{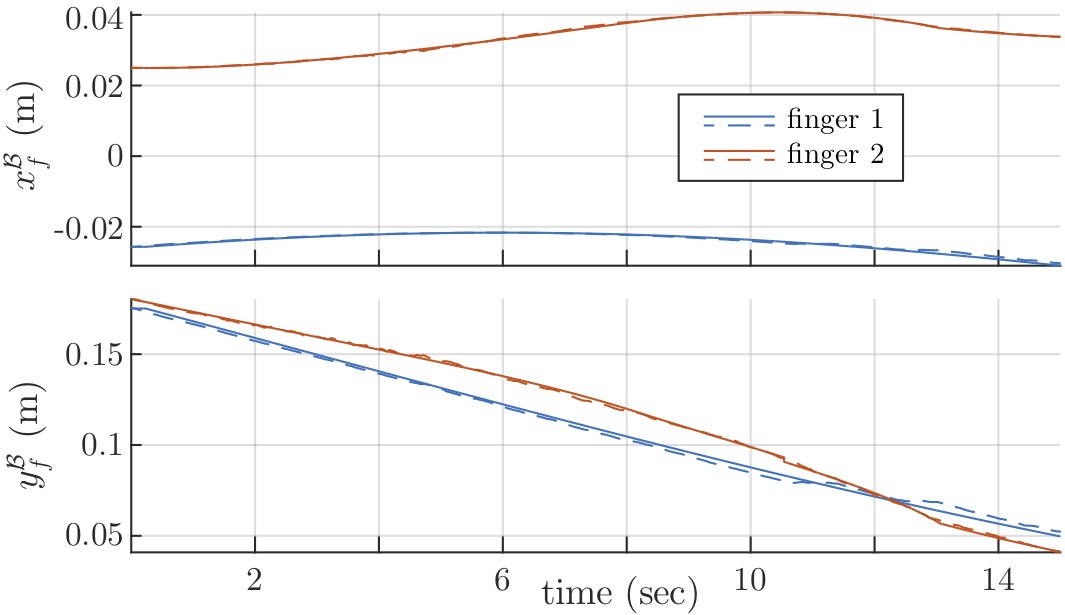}
	\vspace{-0.in}
	\caption{Parameter fitting result of finger contact point position trajectories $\pBf(t)$. Dashed lines are experimental data and solid lines are fitted results.}
	\label{2D_fitting}
\end{figure}

To test our controlled finger stiffnesses $\KHone$ and $\KHtwo$, and to verify our estimate of friction $\mu$ between the fingertips and the object, we collected data from experiments where we manually configured the initial grasp of the object (similar to what is shown in Figure~\ref{fig:2Dconfig}) and commanded the hand to move in the $-y$-direction for $0.15$\,m.  Using the forward contact mechanics from Section~\ref{subsec:contact_mechanics_forward}, and using an SQP solver to adjust our estimates of $\KHone$, $\KHtwo$, and $\mu$ to minimize the sum of the absolute errors between simulated results and 5000 experimentally-measured finger contact positions, we found good agreement between our controlled finger stiffnesses and the experimentally-estimated finger stiffnesses (see Table~\ref{table:param_fitting}).  Figure~\ref{2D_fitting} shows a comparison between experimental results and simulated results with the fitted friction coefficient and finger stiffnesses.

\newcommand{\spc}{\\[-0.5em]}
\begin{table}
\centering
\begin{tabular}{ @{~}l c@{} @{}c@{} }
\hline \hline 
\\[-0.9em] parameters & ~~~~~~initial guess~~~~~~~~~  & estimated \\[-1.em] \\ 
\hline \spc
$\mu$ & 0.24 & 0.2502 \\  \spc
$\KHone$ (N/m) & $\left[ \begin{array}{@{\,}cc@{\,}} 150 & 0 \\ 0 & 100 \end{array} \right]$ & $\left[ \begin{array}{@{\,}cc@{\,}} 152.06 & 0 \\ 0 & 101.1 \end{array} \right]$ \\ \spc  \spc
$\KHtwo$ (N/m) & $\left[ \begin{array}{@{\,}cc@{\,}} 150 & 0 \\ 0 & 100 \end{array} \right]$ & $\left[ \begin{array}{@{\,}cc@{\,}} 150.23 & 0 \\ 0 & 105.94 \end{array} \right]$ \\ \spc  \spc
\hline \hline \spc
\end{tabular}
\caption{Parameter identification result.}
\label{table:param_fitting}
\end{table}

\subsection{Robust Regrasp Planning}

For this regrasp task---where the finger anchors are rigidly attached to the hand, there are two velocity controls for the hand, and the object is stationary---if we know the sliding directions of each fingertip (downward or upward in this example), there exist unique one-to-one mappings between the hand configuration $\ph$, the anchor positions $\paAll$, the fingertip positions $\pfAll$, and the contact forces $\fcAll$.    

To see this, we start by writing the mapping of anchor positions from $\HH$ to $\W$ as
\begin{equation}
	\pa = \ph + \Rh \pHa,
	\label{eq:pa_H2W}
\end{equation}
where $\Rh$ is the rotation matrix of $\HH$. Based on the previous assumptions, $\Rh$ and $\pHa$ are fixed.

When the fingertips slide on the object, each $\fc$ is along an edge of the fingertip's friction cone into the object.  
We denote $\fcperp$ as the direction perpendicular to the contact force $\fc$, so
\begin{equation}
	\fcperp \cdot \fc = 0 \rightarrow \fcperp^\myT \fc = 0.
	\label{eq:fcDot}
\end{equation}
Given a fingertip contact position, the direction $\fcperp$ can be obtained from the object geometry, contact friction, and the sliding direction. 
Substituting Equations~\eqref{eq:fc} and \eqref{eq:pa_H2W} to \eqref{eq:fcDot}, we can solve the hand position for a given pair of finger contact positions $\{\pfOne,\,\pfTwo\}$ as
\begin{equation}
	\ph = \left[ \begin{matrix}
		  \fcperpOne^\myT \KOne \vspace{0.05in} \\ 
	 	  \fcperpTwo^\myT \KTwo
		  \end{matrix}	  \right]^{-1} 
		  \left[  \begin{matrix}
		  {\Delta}_1 \vspace{0.05in} \\ 
	 	  {\Delta}_2
\end{matrix}		   \right],
	\label{eq:ph_2D}
\end{equation}
where ${\Delta}_i = \fcperp^\myT \K (\pf - \Rh \pHa)$. 

Knowing $\ph$, the fingertip contact forces can be solved using Equation~\eqref{eq:fc}. Combined with Equations~\eqref{eq:total_wc} and \eqref{eq:ext_force_bal}, we can test if the fingertip contact wrenches can be balanced by the external contacts.

\subsubsection{Finger Contact Position Map}
\label{subsubsec:FCmap}
\begin{figure}
	\centering
	\includegraphics[width = 3in]{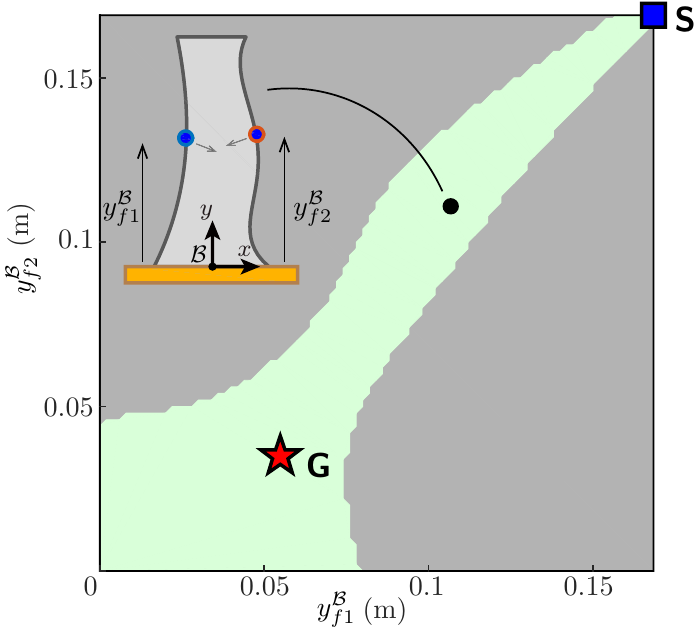}
	\vspace{-0.in}
	\caption{The finger contact position map (FCmap) when each finger moves downward on the object.  Fingertip contact configurations in the green area correspond to wrench balance.  An example configuration is called out, and the direction of each contact force on the boundary of its friction cone is shown.  An example regrasp task is shown by the initial configuration $\BF S$ and the goal configuration $\BF G$.
        }
	\label{fig:feasi_map}
\end{figure}

For the given object, the fingertip contact positions can be parametrized by their $y$-positions in the object frame $\mathcal{B}$.  Figure~\ref{fig:feasi_map} shows the two-dimensional finger contact position map ({FCmap}), with axes defined by $y^{\B}_{f1}$ and $y^{\B}_{f2}$, when both fingers slide downward on the object.  For each point $(y^{\B}_{f1}, y^{\B}_{f2})$ on the FCmap, we can uniquely calculate $\pfAll$, $\paAll$, $\fcAll$, and $\ph$, as described above.  Based on Equations~\eqref{eq:total_wc} and \eqref{eq:ext_force_bal}, we can test if the fingertip forces can be balanced by the external contacts with a linear program:
\begin{equation}
	\underset{{\bar{\bm{\beta}}}}{\text{min}} ~ \mathbf{1}^\myT \bar{\bm{\beta}} , ~~  
	\text{subject to}  \begin{cases}
\What \bar{\bm\beta} = -\bar{\wcAll} - \wg \\ 
\bar{\bm\beta} \geq \mathbf{0}^{p \times 1}
\end{cases}\hspace{-0,15in}.
\label{eq:linprog}
\end{equation}
If a solution $\bar{\bm \beta}$ is found, the fingertip contact locations can satisfy the wrench-balance constraint.
In Figure~\ref{fig:feasi_map}, feasible contact point positions are colored green.

Figure~\ref{fig:feasi_map} also shows an example regrasp task, where $\BF S$ corresponds to the initial fingertip configuration and $\BF G$ corresponds to the goal fingertip configuration.  The regrasp is achievable by fingertips always sliding in the downward direction if and only if $\BF S$ and $\BF G$ are in the same green connected component.

\subsubsection{Planning Algorithm}

Sliding regrasp motion planning is divided into two phases:  Phase 1 ($t \in [0,T_1]$), where the fingertips stick to the object and the anchors are repositioned to bring contact forces to the boundaries of the friction cone, and Phase 2 ($t \in [T_1,T_2]$), where the fingertips slide on the object to the desired new configuration $\BF G$ in the FCmap.
An optional Phase 3 would reposition the anchors again to move the contact forces away from the boundaries of the friction cones.

{\bf \emph{Phase 1, anchor repositioning:}}  The hand trajectory $\ph(t), t \in [0,T_1]$, and therefore the anchor trajectories, is chosen to be a cubic polynomial of time.  This polynomial is uniquely defined by the duration $T_1$, the initial and final velocities $\Dph(0) = \Dph(T_1) =\mathbf{0}$, the initial configuration $\ph(0) = \mathbf{p}_{h0}$, and the final configuration at the point $\BF S$ on the FCmap.  The point $\BF S$ is defined by the fingers' initial contact locations and the fact that the fingers will slide downward, as described above.  $\BF S$ is the unique point of intersection between the space of anchor positions that cause no sliding when the fingertips are at their initial configuration and the space of the FCmap, where the fingers slide downward on the object.

\begin{figure}
	\centering
	\includegraphics{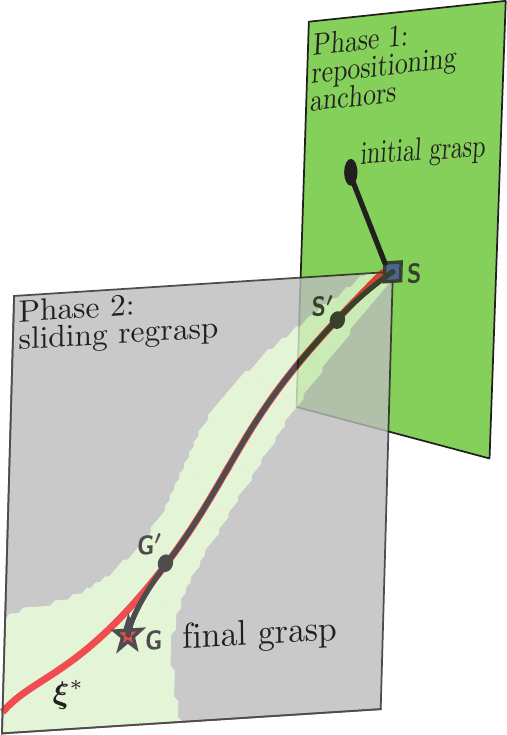}
	\vspace{-0.in}
	\caption{In Phase 1 of the sliding regrasp, the anchors move but the fingertips remain stationary.  At the transition to Phase 2, at the point $\BF S$, the contact forces have moved to the boundary of their friction cones, and the fingertips begin to slide.  Phase 2 is plotted in the FCmap corresponding to both fingers sliding downward on the object.  The fingertips follow the curve of placements $\bm \xi^*$ that maximize $\varepsilon$-robustness (in red) for most of the plan.  The full regrasp plan consists of the hand trajectory $\ph (t)$, $t \in [0,T_2]$, that uniquely corresponds to the curve in black.}
	\label{fig:2stageplan}
\end{figure}

During Phase 1 the hand translates along a straight line with a quadratic velocity profile beginning and ending at rest.  Fingertip forces
are guaranteed to remain within their respective friction cones during the straight-line motions of the anchors due to the convexity of the friction cones.
Figure~\ref{fig:2stageplan} gives a conceptual representation of the hand's motion during Phase 1, which ends when the anchors have moved so that the grasp configuration is at $\BF S$, which resides in both the FCmap and the space of anchor configurations that does not cause sliding at the fingertips.

{\bf \emph{Phase 2, sliding regrasp:}} 
Since the fingertip contact positions can be described by the coordinates $(y^{\B}_{f1}, y^{\B}_{f2})$, we use $\bm\xi(t) = [y^{\B}_{f1}(t),\; y^{\B}_{f2}(t) ]^\myT, t \in [T_1,T_2]$, to represent sliding trajectories. To accomplish the desired regrasp we have $\bm\xi(T_1) = \BF S$ and $\bm\xi(T_2) = \BF G$. 

A sliding trajectory $\bm\xi(t)$ is feasible if it always lies in the feasible region of FCmap. 
Based on the findings in Section~\ref{sec:robustness}, the further away the required external contact wrench $\bar{\mathbf{w}}_e$ is from the boundaries of $\WCe$, the more robust a fingertip configuration is. 
For this task, given the contact position of one finger, there is an optimally robust contact position of the other finger. The union of these most robust fingertip position pairs is a curve in the FCmap, denoted $\bm\xi^*$. 
To describe how far a wrench $\weAll$ is from the faces of the wrench cone $\WCe$, we define a matrix $\Wperp $
 whose rows are unit vectors normal to the faces of $\WCe$ and pointing into the cone. The curve $\bm\xi^*$ is found by the following procedure:
\[
	\forall \,y^{\B}_{f1},~\text{find} ~ y^{\B*}_{f2} ~\text{maximizing } d \text{ such that }
	\Wperp \bar{\mathbf{w}}_e \geq d,
\]
where 
$\bar{\mathbf{w}}_e$ is the total expected external contact wrench.
The solved $\bm\xi^*$ is shown as the red curve in Figure~\ref{fig:2stageplan},
consisting of points calculated at 1~mm increments in $y_{f1}^{\mathcal{B}}$.  The entire FCmap as shown in Figures~\ref{fig:feasi_map} and \ref{fig:2stageplan} is not explicitly computed during planning; it is only shown to help visualize the planning space and to illustrate the notion of robustness.

To maximize robustness, the principle of our planning algorithm is to plan $\bm\xi(t)$ to coincide with $\bm\xi^*$ as much as possible while satisfying the desired final regrasp. By introducing a point $\BF S'$ where $\bm\xi(t)$ reaches $\bm\xi^*$ from $\BF S$, and a point $\BF G'$ where $\bm\xi(t)$ departs $\bm\xi^*$ to go to $\BF G$, the sliding trajectory $\bm\xi(t)$ is defined by three pieces:
\begin{itemize}
	\item \emph{1st piece}\,($\BF S \rightarrow \BF S'$, $T_1 \leq t \leq T_{21} = T_1 + \Delta T_{21}$)\,: The contact sliding trajectories $\bm\xi(t)$ are cubic time polynomials of duration $\Delta T_{21}$, solved uniquely by the four boundary conditions $\bm\xi(T_1) = \BF S$, $\bm\xi(T_{21}) = \BF S'$, $\dot{\bm\xi}(T_1) = \mathbf{0}$, and $\dot{\bm\xi}(T_{21}) = \mathbf{v}_{s}$, where $\mathbf{v}_s$ is determined by the initial velocity of the next piece.
	\item \emph{2nd piece}\,($\BF S' \rightarrow \BF G'$, $T_{21} \leq t \leq T_{22} = T_{21} + \Delta T_{22}$)\,: The contacts slide along $\bm\xi^*$ for a duration $\Delta T_{22}$. The sliding velocities are assumed to have a constant magnitude $\|\dot{\bm\xi}\| = v_2 = L_2/\Delta T_{22}$, where $L_2$ is the arclength of $\bm\xi^*$ between $\BF S'$ and $\BF G'$. The initial and final velocities are $\mathbf{v}_s = v_2 \, \hat{\partial \bm\xi^*} |_{\BF S'} $ and $\mathbf{v}_g = v_2 \, \hat{\partial \bm\xi^*} |_{\BF G'} $, where $\hat{\partial \bm\xi^*} |_{\BF X}$ is the normalized tangent vector at point ${\BF X}$.
	\item \emph{3rd piece}\,($\BF G' \rightarrow \BF G$, $T_{22} \leq t \leq T_2 = T_{22} + \Delta T_{23}$)\,: The contacts slide from $\BF G'$ to $\BF G$ following cubic time polynomials of duration $\Delta T_{23}$, solved uniquely by the four boundary conditions $\bm\xi(T_{22}) = \BF G'$, $\bm\xi(T_{2}) = \BF G$, $\dot{\bm\xi}(T_{22}) = \mathbf{v}_g$, and $\dot{\bm\xi}(T_{2}) = \mathbf{0}$.
\end{itemize}

The design variables for Phase 2 are the via points $\BF S'$ and $\BF G'$ on $\bm \xi^*$ and the durations $\Delta T_{21}$, $\Delta T_{22}$, and $\Delta T_{23}$.  The objective function can be expressed as maximizing a function of robustness (e.g., how much the planned sliding trajectory coincides with $\bm \xi^*$) while penalizing large sliding velocities.  One formulation of the motion planning problem is the following nonlinear program:
\begin{align}
  \textbf{find} & \quad \BF S', \BF G', \Delta T_{21}, \Delta T_{22},\Delta T_{23} \nonumber \\
	\textbf{maximizing} & \quad L_2(\bm\xi^*, \BF S', \BF G') - \kappa V_{\text{max}} \nonumber \\
	\textbf{such that} & \quad \text{1) } \texttt{sgn}(\dot{\bm \xi}) = \texttt{sgn}(\BF G-\BF S) \nonumber \\
        & \quad \text{2) } \Delta T_{21} + \Delta T_{22} + \Delta T_{23} = T_2 - T_1,  \nonumber
\end{align}
where $\kappa$ is a positive weighting scalar and $V_{\text{max}} = \text{max}_t(|\dot{y}_{f1}^{\mathcal{B}}(t)| + |\dot{y}_{f2}^{\mathcal{B}}(t)|)$.
The first constraint ensures that the sliding directions are always towards the goal, as assumed in Section~\ref{subsubsec:FCmap}.

\begin{figure*}[t]
	\centering
	\includegraphics[width = 7.1in]{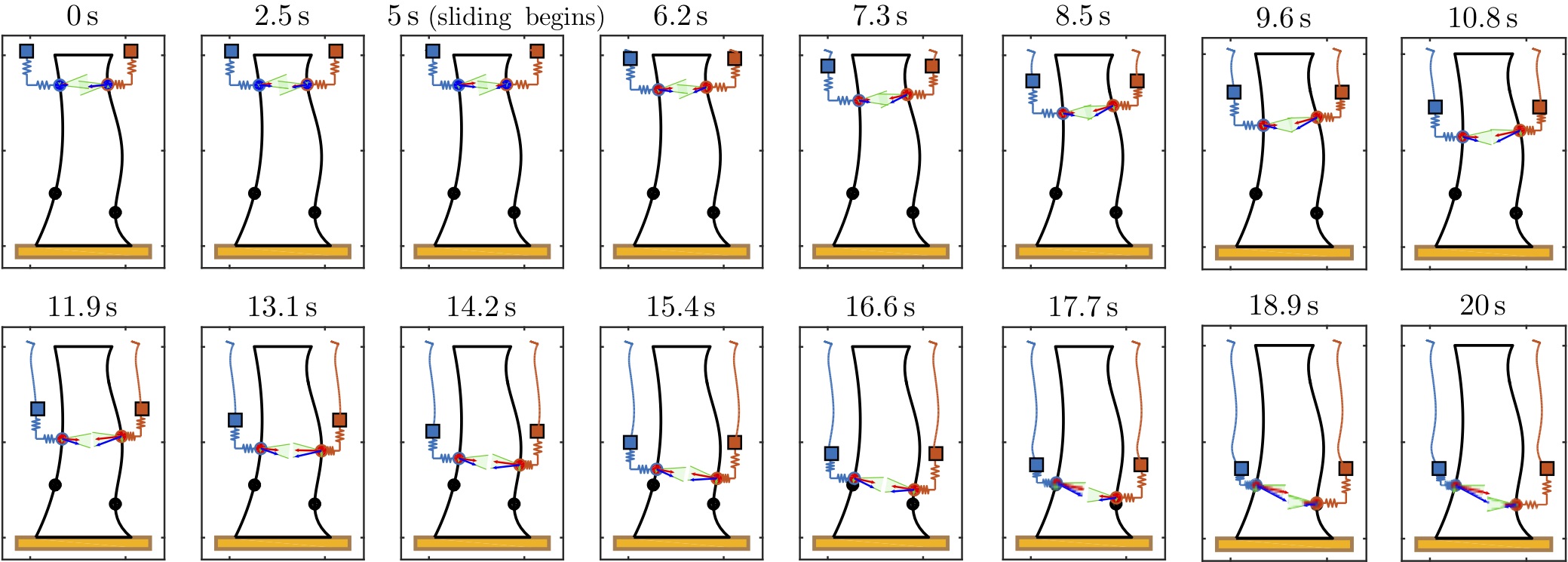}
	\vspace{-0.in}
	\caption{Snapshots of the planned motion. Positions are in meters (0.1\,m/div).
	Small squares show the finger anchors. Blue (sticking) and red (sliding) dots show finger contact points from simulation. Black dots are the goal contact positions. 
	Green lines shows the edges of the contact friction cones. Blue, red, and green arrows (only visible by zooming in) show contact forces, contact normal forces, and contact tangential forces respectively. }
	\label{fig:plan_motion}
      \end{figure*}

\subsubsection{Experimental Results}
We defined a sliding regrasp task by $\BF S = [0.168~\text{m}, 0.169~\text{m}]^\myT$ and $\BF G = [0.055~\text{m}, 0.035~\text{m}]^\myT$, where the initial configuration of the hand is such that the fingertip contact forces are in the interior of the friction cone.  Given $T_1 = 5~\text{s}$, $T_2 = 20~\text{s}$, and $\kappa = 0.5$, and using MATLAB's {\tt fmincon}, we find the Phase 2 sliding regrasp plan shown as the black curve in Figure~\ref{fig:2stageplan}.  As expected, the curve $\bm \xi(t)$ coincides with the optimally robust curve $\bm\xi^*$ for much of the Phase 2 portion of the plan, to maximize robustness to force disturbances.  The full plan, showing the repositioning of the hand (and anchors) for $5$~s in Phase 1 and the Phase 2 sliding for $15$~s, is shown in snapshots in Figure~\ref{fig:plan_motion}.

\begin{figure}
	\centering
	\includegraphics[width = 3.4in]{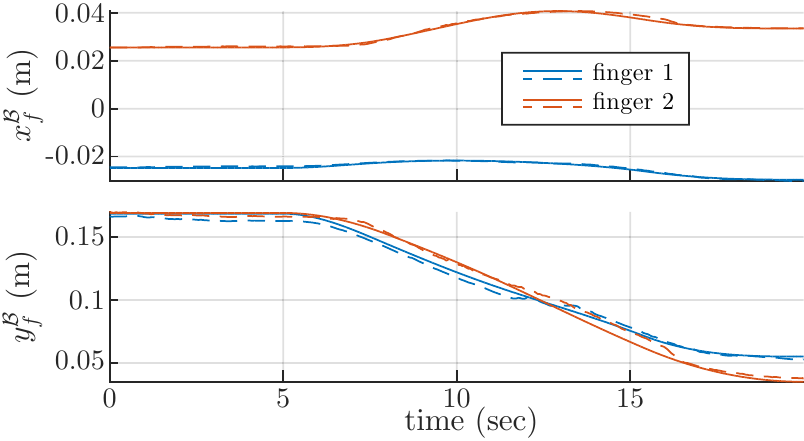}
	\vspace{-0.in}
	\caption{An example experimental result showing contact point positions in $\mathcal{B}$. Dashed lines are experimental data and solid lines are simulated trajectories.}
	\label{fig:exp_res}
\end{figure}

Experimental implementations of the plan followed the expected motions closely, indicating that the robustness-maximizing regrasp planner does indeed deliver a robust motion plan.  During execution of the sliding regrasp, the hand's motion was feedback-controlled to follow the planned hand trajectory, and the stiffnesses of the fingertips were actively controlled.
The fingertips were not individually motion-controlled to try to track the planned fingertip trajectories.  Figure~\ref{fig:exp_res} shows a typical experimental result compared to the planned regrasp.  The final fingertip positions deviated from the planned positions by $2.2~\text{mm}$ and $2.6~\text{mm}$ for fingers one and two, respectively, compared to total travel distances of $114.2~\text{mm}$ and $136.3~\text{mm}$.

\section{Discussion and Future Work}%
\label{sec:conclusion}%

In this paper we introduced the concept of spring-sliding compliance for in-hand sliding regrasp by pushing the grasped object against environmental constraints.  Sliding provides a passive mechanical nonlinear velocity ``compliance'' to tangential forces, and spring compliance maintains contact normal forces as the fingertips slide over the object.  Spring compliance achieves contact normal force control by motion control of physical or virtual finger anchors.  We derived the finger contact forward and inverse mechanics for spring-sliding compliant contacts and formulated the $\varepsilon$-robustness condition for sliding regrasps.  An experimental implementation of the theory on a two-fingered robot hand shows that spring-sliding regrasps can be automatically planned and robustly executed.

Future work may include modifying the point fingertips, to allow fingertips of more general geometry, and patch contacts, with their ability to provide friction forces resisting spin about fingertip contact normals.  This increases the complexity of the analysis and, in the most general case, would require modeling fingertip compliance as a $6 \times 6$ matrix, including three rotational freedoms.  These more complex models may be justified by better robot hands that reliably control contact compliance and sense contact locations and forces.

In this paper we specified the environmental contact locations and finger contact mode sequences. In future work the motion planning algorithm could be expanded to judiciously choose the environmental contacts and sequences of fingertip sticking and sliding phases to add more design freedoms.
Also, while we focused on stationary contact between the object and the environment, spring-sliding regrasps could be obtained with sliding or rolling contacts with the environment, even allowing tasks that assemble the object with the environment.  For sliding regrasps with moving contacts with the environment, feedback control (not considered in this paper) could be employed to stabilize plans that do not meet the restrictive definition of $\varepsilon$-robustness.

Finally, learning methods could be employed to account for unmodeled effects beyond contact force uncertainty.  The modeling in this paper can serve to bootstrap learning, allowing more efficient use of data obtained from experiments and learning of corrections to the model rather than learning from scratch.

\appendices

\section{Compliant Grasps via Open-Loop Torque-Controlled Joints}
\label{app:torque}

Passively compliant grasps may arise from fingers under open-loop joint-torque control (e.g., constant torques or currents at the joints).  As one example, assume the world frame is at the finger base and $\mathbf{p}_f$ is the fingertip position relative to the anchor. Let $\bm{\theta}$ denote the finger joint angle vector, $\bm\uptau$ denote the joint torque vector, and $\mathbf{J}(\bm\theta)$ denote the Jacobian matrix sastisfying $\dot{\mathbf{p}}_f = \mathbf{J}\dot{\bm \theta}$. From finger kinematics and the principle of virtual work, we have the mapping from fingertip contact forces to the joint torques $\bm\uptau = \mathbf{J}^\myT \mathbf f_c\,$. When $\mathbf J$ is invertible, we have
\begin{align}
	\mathbf f_c &= \mathbf{J}^{-\myT} \bm\uptau \nonumber\\
	\rightarrow \partial \mathbf{f}_c &= \partial(\mathbf{J}^{-\myT}) \bm\uptau + \mathbf{J}^{-\myT} \partial\bm\uptau.
	\label{eq:partial_fc}
\end{align}
From the definition of the Jacobian we have
\begin{equation}
	\partial \mathbf{p}_f = \mathbf{J} \partial\bm\theta.
	\label{eq:partial_pf}
\end{equation}
Combining Equations \eqref{eq:partial_fc} and \eqref{eq:partial_pf}, we can write the finger stiffness matrix as
\newcommand{\tK}{\mathbf{K}}
\begin{equation}
	\tK = -\frac{\partial \mathbf{f}_c}{\partial \mathbf{p}_f} = -\frac{\partial(\mathbf{J}^{-\myT})}{\partial\bm\theta} \bm\uptau \mathbf{J}^{-1} -  \mathbf{J}^{-\myT} \frac{\partial \bm\uptau}{\partial\bm\theta} \mathbf{J}^{-1}.
	\label{eq:local_stiffness}
\end{equation}
The specific expression for $\tK$ depends on the Jacobian and the joint torques $\bm\uptau$.

Continuing the example, assume that joint torques are independent of the finger position ($\frac{\partial\bm\uptau}{\partial \bm\theta} = \mathbf 0$) for the
two-joint
finger shown in Figure~\ref{fig:finger_springmodel}(b). Assume that the links have unit length and the joint torques have a constant value of $1$. Then the stiffness matrix in Equation~\eqref{eq:local_stiffness} simplifies to
\begin{equation}
	\tK(\bm{\theta}) = -\frac{\partial(\mathbf{J}^{-\myT})}{\partial\bm\theta} \left[\begin{array}{c}1\\1\end{array}\right]	 \mathbf{J}^{-1} = \left[ \begin{array}{cc}
		k_{11} & k_{12}\\
		k_{21} & k_{22}
	\end{array}\right],
	\label{eq:2R_K}
\end{equation}
where 
\begin{align*} 
k_{11} = &\frac{1}{4} \csc ^3\theta _2 \left(\cos \left(2 \theta _1-\theta _2\right)+ \right. \\
&2 \left(\cos \theta _2+ \cos \left(2 \theta _1+2\theta _2\right)+1\right) 
\left.+\cos \left(2 \theta _1+\theta _2\right)\right), \\
k_{12} = &~k_{21} = \frac{1}{4} \left(\sin \left(2 \theta _1-\theta _2\right)+2 \sin \left(2 \theta _1+2\theta _2\right) + \right. \\
&\left. \sin \left(2 \theta _1+\theta _2\right)\right) \csc ^3\theta_2, \\
k_{22} = &-\frac{1}{4} \csc ^3\theta _2 \left(\cos \left(2 \theta _1-\theta _2\right)- \right. \\
&\left. 2 ( \cos \theta_2 - \cos \left(2 \theta _1+2\theta _2\right) + 1) +\cos \left(2 \theta _1+\theta _2\right)\right) .
\end{align*}
The eigenvalues of the stiffness matrix $\tK$ are
\begin{gather*}
	\uplambda_1 = \frac{1}{2} \csc ^3\theta _2 \left(1 + \cos \theta_2 - \sqrt{1 + \cos \left(3 \theta _2\right) + \cos \theta_2 + \cos^2\theta _2}\right) , \\
	\uplambda_2 = \frac{1}{2} \csc ^3\theta _2 \left(1 + \cos \theta_2 + \sqrt{1 + \cos \left(3 \theta _2\right) + \cos \theta_2 + \cos^2\theta _2}\right).
\end{gather*}

\begin{figure}[t]
  \begin{center}
    \frame{\includegraphics[width=3.35in]{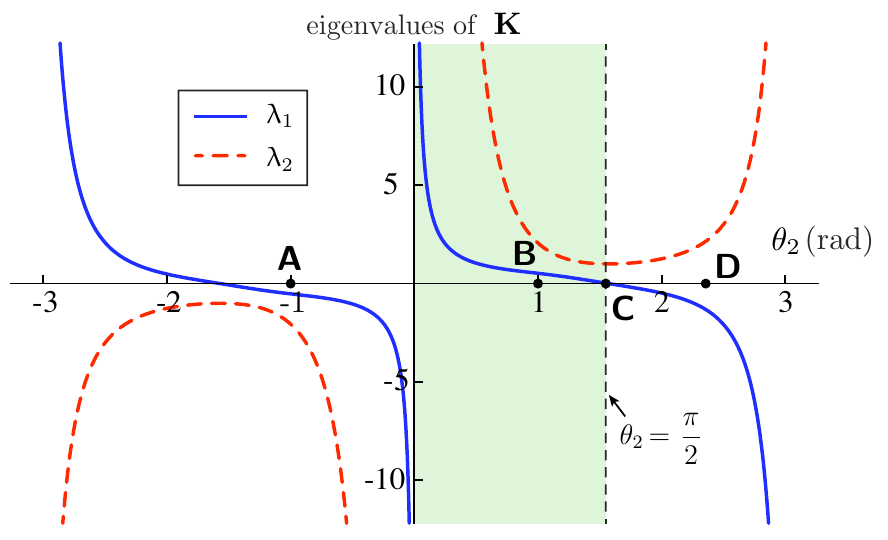}}
        \frame{\includegraphics[width=3.35in]{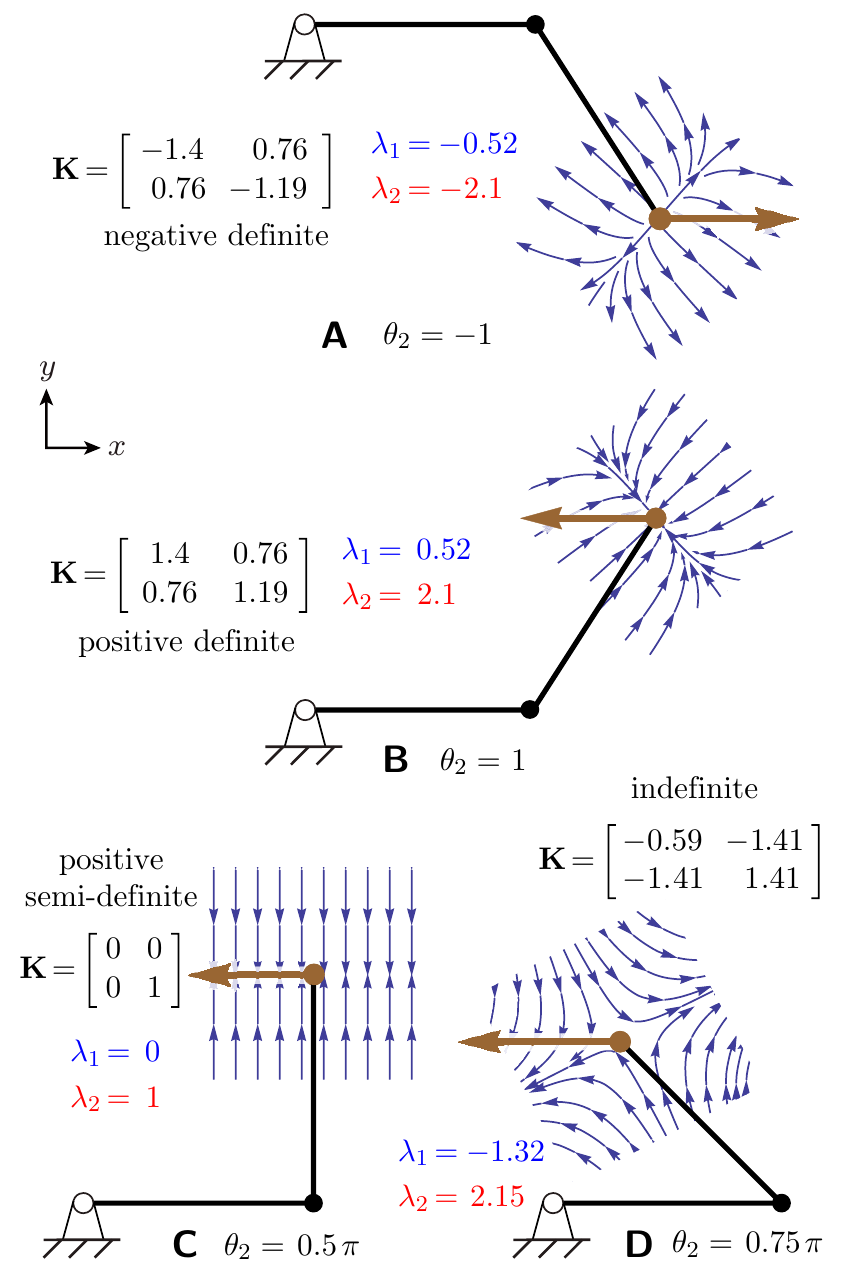}}
        \end{center}
	\caption{(Top) Eigenvalues of the stiffness matrix $\mathbf{K}$ for the example 2R finger with $\uptau_1=\uptau_2=1\,\text{Nm}$ and link lengths of $1\,$m.  The green shaded region shows the range of $\theta_2$ that yields positive-definite $\mathbf{K}$.
          (Bottom) Illustrations of the finger configurations \textbf{\textsf{A}}, \textbf{\textsf{B}}, \textbf{\textsf{C}}, and \textbf{\textsf{D}} with $\theta_1=0$.   Each chosen $\theta_2$, and the corresponding stiffness matrix $\tK$ and its eigenvalues by Equation~\eqref{eq:2R_K}, are shown.     The stiffness is visualized as streamplots: for small fingertip location virtual displacements $\partial \mathbf{p}_f$ relative to the current fingertip location $\mathbf{p}_f$, the net change in the force at the fingertip $\partial \mathbf{f}_c$ (due to the joint torques) is in the direction of the arrows shown on the streamplot.  Brown arrows represent the fingertip force $\mathbf{f}_c$ at the nominal configuration. }
	\label{fig:2R_eigenvalues}
\end{figure}

The eigenvalues are only related to $\theta_2$ since $\theta_1$ only changes the finger's orientation relative to the base. 
The stiffness matrix $\tK$ is symmetric and the two eigenvalues must both be positive to satisfy the assumption of positive-definite stiffness. We plot the eigenvalues with respect to $\theta_2$ in Figure~\ref{fig:2R_eigenvalues} (Top). 
Figure~\ref{fig:2R_eigenvalues} (Bottom) shows the finger configuration and stiffness for four values of $\theta_2$.
The finger configuration should satisfy $0<\theta_2<\frac{\pi}{2}$ to satisfy the positive-definite stiffness assumption of this paper. In cases \textbf{\textsf{A}}, \textbf{\textsf{C}}, and \textbf{\textsf{D}}, the stiffness matrix is not positive definite, which may lead to ``runaway'' sliding where the quasistatic condition is violated.  

\begin{figure}
	\centering
	\includegraphics[width=3.in]{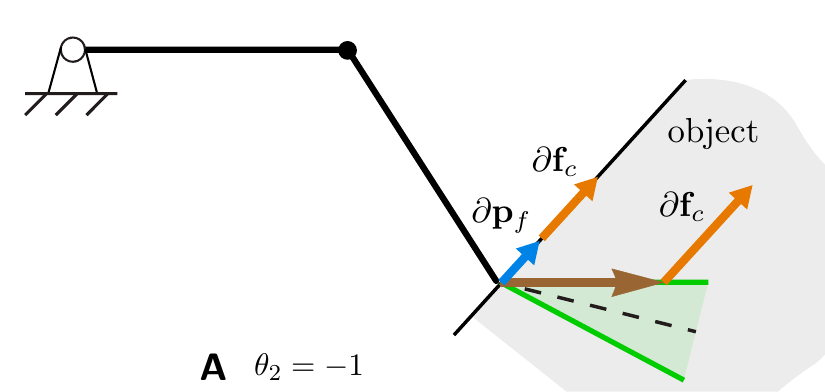}
		\caption{An unstable sliding example for case \textbf{\textsf{A}} in Figure~\ref{fig:2R_eigenvalues}: since $\uptau_1=\uptau_2$ the fingertip force is always aligned with the first link. For a fingertip displacement $\partial \mathbf{p}_f$ shown as the blue vector, the force applied by the joints at the fingertip changes as shown by $\partial \mathbf{f}_c$.   The green shaded area is the friction cone.}
	\label{fig:2R_ill_sliding}
\end{figure}

As an example, Figure~\ref{fig:2R_ill_sliding} shows case \textbf{\textsf{A}} of Figure~\ref{fig:2R_eigenvalues}.  Since $\uptau_1=\uptau_2$, the fingertip force is always aligned with the first link of the finger.  For the friction cone shown, the contact force with the stationary object is initially on the edge of the friction cone and the finger is force balanced.  If the contact location on the object is perturbed by $\partial \mathbf{p}_f$, as shown, the change $\partial\mathbf{f}_c$ in the fingertip force generated by the joint torques causes the total force to move outside the friction cone, meaning friction forces applied by the object to the finger can no longer completely balance the finger force.  The fingertip will accelerate in the sliding direction and the motion of the fingertip must be solved for using dynamics; the quasistatic equilibrium assumption is violated.  Conditions where the quasistatic assumption are violated are studied further in Section~\ref{subsec:sliding_ill_condition}. 

In summary, many models of the finger hardware and control strategy satisfy the assumptions of this paper, even certain configurations of the simple open-loop torque-controlled fingers described above.

\bibliographystyle{IEEEtran}
\bibliography{InhandSSCBibtex}

\end{document}